%% file: neu.tex
\documentclass{article}
\usepackage[final]{neurips_2023}

\usepackage[utf8]{inputenc} % allow utf-8 input
\usepackage[T1]{fontenc}    % use 8-bit T1 fonts
\usepackage{hyperref}       % hyperlinks
\usepackage{url}            % simple URL typesetting
\usepackage{booktabs}       % professional-quality tables
\usepackage{amsfonts}       % blackboard math symbols
\usepackage{nicefrac}       % compact symbols for 1/2, etc.
\usepackage{microtype}      % microtypography
\usepackage{xcolor}         % colors

\title{PuzzleFusion: Unleashing the Power of\\ Diffusion Models for Spatial Puzzle Solving}

\author{Sepidehsadat Hosseini, Mohammad Amin Shabani, Saghar Irandoust\thanks{Work done while being student at SFU} , Yasutaka Furukawa\\
Simon Fraser University\\
{\text{\{sepidh, mshabani, sirandou, furukawa\}@sfu.ca}}}
\newcommand{\mysubsubsection}[1]{\vspace{0.1cm} \noindent {\bf #1}:}

\newcommand{\mypara}[1]{\vspace{0.1cm} \noindent {\bf #1}}

\makeatletter
\DeclareRobustCommand\onedot{\futurelet\@let@token\@onedot}
\def\@onedot{\ifx\@let@token.\else.\null\fi\xspace}

\def\eg{\emph{e.g}\onedot} 
\def\ie{\emph{i.e}\onedot}

\def\etal{\emph{et al}\onedot}
\makeatother

\usepackage{array}
\usepackage{graphicx}
\usepackage{amsmath}
\usepackage{amssymb}
\usepackage{multirow}
\usepackage{capt-of}
\usepackage{epsfig}
\usepackage{comment}
\usepackage{float}
\usepackage{booktabs}
\usepackage[ruled,vlined]{algorithm2e}
\usepackage{pifont}% http://ctan.org/pkg/pifont
\newcommand{\xmark}{\ding{55}}%
\usepackage{wrapfig}

\usepackage[capitalize]{cleveref}

\crefname{section}{Sec.}{Secs.}
\Crefname{section}{Section}{Sections}
\Crefname{table}{Table}{Tables}
\crefname{table}{Tab.}{Tabs.}

\begin{document}

\maketitle

% {
%  \maketitle
%  \vspace{-2em}
%  \centerline{
%  \includegraphics[width=\textwidth]{figures/tea2.pdf} 
% }
% \captionof{figure}{Extreme Structure from Motion is the task of taking
% a set of room layouts and their corresponding room types as the input and predicting the position and the orientation of each room. The biggest discovery and surprise of this paper is that conditional generation by a Diffusion Model solves this challenging pose estimation problem.}
% %refines each model by distilling information from neighboring %clusters. }
% \label{fig:teaser}
% \vspace{1em}
% }
\begin{abstract}
   \input{sections/0_abstract} 
\end{abstract}

%%%%%%%%%%%%%%%%%%%%%%%%%%%%%%%%%%%
\input{sections/1_intro}

\input{sections/2_related_work}
\begin{figure}[!tbh]
   \centering
\includegraphics[width=0.975\textwidth]{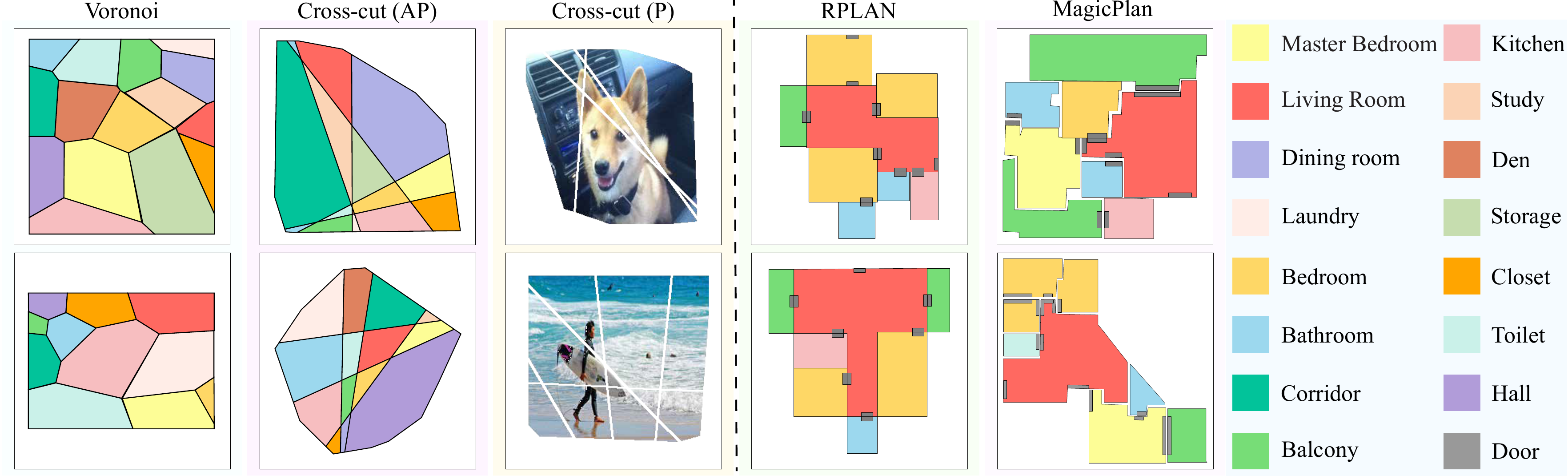}
    \caption{Spatial puzzle datasets for Voronoi and Cross-cut puzzle solving (Left) and room layout arrangement (Right). We consider both pictorial and apictorial versions of the Cross-cut jigsaw puzzle. Note that colors in puzzle solving problems are random and just indicate different pieces.}
    %The left three show samples for the three jigsaw puzzle tasks. 
    %MagicPlan is a new dataset offered by MagicPlan from its production pipeline. }
    %RPLAN~\cite{rplan_citation} is an existing synthetic floorplan dataset.}
    \label{fig:dataset}
\end{figure}
 \input{sections/3_dataset}

\input{sections/4_preliminaries}

 \input{sections/5_method}
 \input{sections/7_experiments}

 \input{sections/8_conclusion}
\mysubsubsection{Acknowledgment}
This research is partially supported by NSERC Discovery Grants with Accelerator Supplements and the DND/NSERC Discovery Grant Supplement, NSERC Alliance Grants, and the John R. Evans Leaders Fund (JELF). We are also thankful to Magicplan for sharing the datasets.
%%%%%%%%%%%%%%%%%%%%%%%%%%%%%%%%%%%

\bibliography{egbib}

 \appendix
% % The supplementary document provides more details on our system and the competing methods (Sect.~\ref{sec:arc}), more details on JigsawPlan and RPLAN datasets (Sect.~\ref{sec:dataset}), ablation studies on RPLAN dataset (Sect.~\ref{sec:abs}), and additional qualitative examples  (Figs.~\ref{fig:main}, \ref{fig:steps}, and \ref{fig:s-f}) as promised in the main paper. Figure~\ref{fig:main} shows more qualitative evaluations of our approach against the three competing methods. Figure~\ref{fig:steps} visualizes the samples predicted by our method at step t. Figure~\ref{fig:s-f} shows more qualitative evaluations of our method for Full JigsawPlan and Full RPLAN datasets.

% The supplementary document provides more details on our system and the competing methods (Sect.~\ref{sec:arc}), more details on the datasets (Sect.~\ref{sec:dataset}), additional ablation studies (Sect.~\ref{sec:abs}), and additional qualitative examples  (Figs.~\ref{fig:steps}, and \ref{fig:s-f}, \ref{fig:main}, \ref{fig:qualitative_voronoi}, \ref{fig:qualitative_crosscut}) as promised in the main paper. Figure~\ref{fig:steps} visualizes the samples predicted by our method at step t. Figure~\ref{fig:s-f} shows more qualitative evaluations of our method for Full MagicPlan and Full RPLAN datasets. Figure~\ref{fig:main} shows more qualitative evaluations of our approach against the three competing methods. 
% %
% Please also see the supplementary video for more examples.

\input{sections/sups-method}

\input{sections/sups-dataset}

\input{sections/sups-exps}

%%%%%%%%%%%%%%%%%%%%%%%%%%%%%%%%%%%%%%%%%%%%%%%%%%%%%%%%%%%%

\end{document}

%% file: sections/0_abstract.tex
This paper presents an end-to-end neural architecture based on Diffusion Models for spatial puzzle solving, particularly jigsaw puzzle and room arrangement tasks.
In the latter task, for instance, the proposed system % ``PuzzleFusion'' 
takes a set of room layouts as polygonal curves in the top-down view and aligns the room layout pieces by estimating their 2D translations and rotations, akin to solving the jigsaw puzzle of room layouts. A surprising discovery of the paper is that the simple use of a Diffusion Model effectively solves these challenging spatial puzzle tasks as a conditional generation process. 
To enable learning of an end-to-end neural system, the paper introduces new datasets with ground-truth arrangements: 1) 2D Voronoi jigsaw dataset, a synthetic one where pieces are generated by Voronoi diagram of 2D pointset; and 2) MagicPlan dataset, a real one offered by MagicPlan from its production pipeline, where pieces are room layouts constructed by augmented reality App by real-estate consumers.
The qualitative and quantitative evaluations demonstrate that our approach outperforms the competing methods by significant margins in all the tasks.
We will publicly share all our code and data.

%fuses two existing spatial puzzle problems into a unified task, while also introducing more challenging 

%The paper presents two new spatial puzzle tasks with data and ground-truth labels: 1) Voronoi Jigsaw Puzzle where pieces are generated by voronoi diagram of random 2D points, making it more challenging than traditional Crosscut Jigsaw Puzzle where pieces are cut only by lines; and 2) Room arrangement with floorplans for 98,780 houses from a production pipeline (... \yasu{emphasize that this is a real task...}
%
%a new dataset of room layouts and floorplans for 98,780 houses. 
%The qualitative and quantitative evaluations demonstrate that the proposed approach outperforms the competing methods by significant margins. 

%A project website with supplementary video and document is \href{https://sepidsh.github.io/JigsawPlan/index.html}{here}.

%% file: sections/1_intro.tex
\section{Introduction}
Spatial puzzle solving demands meticulous reasoning and arrangement of elements within a given space. The classic example of jigsaw puzzles, which many of us have enjoyed as a recreational activity, showcases the challenge and satisfaction of such tasks. Applications of jigsaw puzzles extend beyond mere entertainment, encompassing areas such as the restoration of shattered 2D artwork and documents~\cite{shed1, shed2}, image stitching in computer vision~\cite{stitch}, and even DNA sequence assembly in genomics~\cite{dna, dna2}. In real estate, room layout arrangement has emerged as a compelling spatial puzzle task, where consumers leverage mobile devices to scan individual rooms that are to be arranged into a floorplan.

Spatial puzzle solving poses a considerable challenge even for humans, making it an engaging mental exercise. Addressing this challenge for computational approaches is even more demanding, despite the emergence of deep learning. Current state-of-the-art techniques typically enumerate pairs of aligned pieces, evaluate their compatibility by learning, and employ optimization or search methods to identify the most likely global arrangement~\cite{harel2021crossing, puzz1,shih2018divide}. However, these approaches struggle to scale as the complexity of global arrangement increases exponentially with the number of pieces.
%, rendering it difficult to harness the full power of advanced learning techniques.

This paper makes a breakthrough in spatial puzzle solving with an end-to-end neural architecture based on Diffusion Models~\cite{denoising}, named PuzzleFusion.  The surprising discovery of this paper is that a Diffusion Model, typically regarded as a powerful generative model, is
%proves to be 
an effective spatial puzzle solver. PuzzleFusion formulates the spatial puzzle as a conditional generation process where the piece information is the condition.
%, making significant boost in accuracy and speed.
%is significantly more accurate and faster than existing methods. 

Concretely, each puzzle piece is a polygonal shape, represented by a sequence of 2D corner coordinates. A piece is associated with task-dependent properties, such as a texture image for pictorial jigsaw puzzles or the room types and door locations for Room Layout Arrangement. The output is a 2D position and an orientation for each piece. The forward diffusion process adds noise to a feature vector initialized with the ground truth position and orientation. The reverse denoising process learns to infer position and orientation subject to the piece shape and property information as a condition.

%We train a diffusion model that generates the 2D translation and the orientation of each piece, subject to the input shape information as a condition.

%4-fold rotation (under the Manhattan assumption) for each room while specifying the input data as the condition, akin to solving a jigsaw puzzle of room layouts.
%our input consists of polygonal shapes (\yasu{correct?} : \sepid{yes although no window only doors so i am taking out window} ) with appearance information for the pictorial jigsaw puzzle task or room types and door locations for the room layout arrangement task). The output is the 2D translation and orientation of each piece. 
%We train a diffusion model to denoise the 2D translation and rotation information, initialized by the Gaussian distribution.
%(either a 4-fold rotation under the Manhattan assumption or a specific rotation angle).

%We evaluated the proposed approach against state-of-the-art jigsaw puzzle and room layout arrangement solvers on the three tasks (cross-cut jigsaw puzzle, Voronoi jigsaw puzzle, and room layout arrangement tasks). To enable robust learning, the paper also introduce large-scale datasets with ground-truth labels for the latter two tasks. In particular, we use a real production data from MagicPlan~\cite{magicplan_company} for the room layout arrangement task, where real-estate consumers use an augmented reality app to semi-interactively build a room layout (shape) that are to be aligned usually fully manually, but we seek to make this automatic.
Qualitative and quantitative evaluations over the three tasks (Cross-cut jigsaw, Voronoi jigsaw, and room layout arrangement) demonstrate that PuzzleFusion outperforms the current state-of-the-art methods by significant margins.
%We evaluated PuzzleFusion against the current state-of-the-art over three tasks: cross-cut jigsaw, Voronoi jigsaw, and room layout arrangement tasks. 
The paper also makes dataset contributions by introducing large-scale datasets with ground-truth labels for Voronoi jigsaw and room layout arrangement tasks.
%, enabling robust learning. 
Specifically, the room layout arrangement dataset comes from a production pipeline by MagicPlan (https://www.magicplan.app/)
%~\cite{magi}, 
consisting of room-layouts and floorplans for 98,780 houses, which we obtained permission to share with the research community.

In summary, this paper makes the following three key contributions: 1) An end-to-end neural architecture based on Diffusion Models for spatial puzzle solving; 2) State-of-the-art performance across three spatial puzzle tasks in terms of accuracy and speed; and 3) The new spatial puzzle datasets including room-layouts and floorplans for 98,780 houses from a production pipeline.
We will make all our code and data public.
\begin{figure*}[!t]
    \centering
    \includegraphics[width=\textwidth]{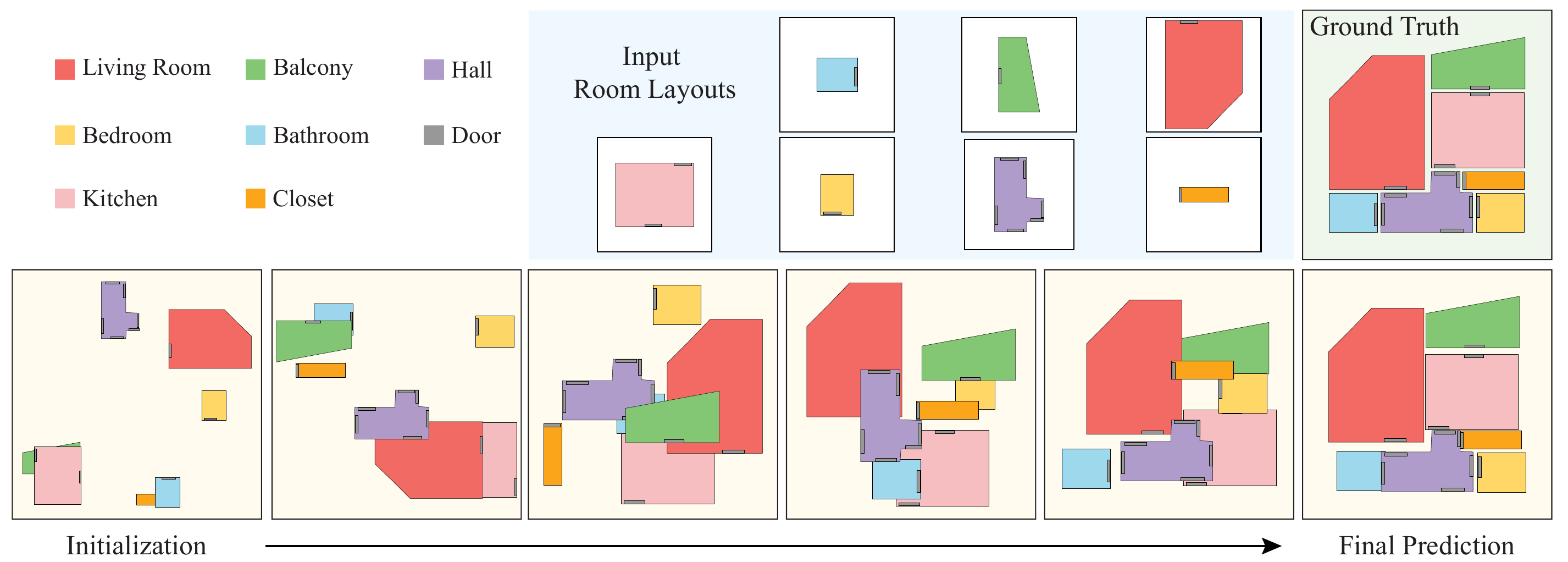} 
    \caption{Room layout arrangement is the task of taking a set of room layouts and their corresponding room types as the input and predicting the position and the orientation of each room. The biggest discovery and surprise of this paper is that conditional generation by a Diffusion Model solves this challenging problem.}
    \label{fig:teaser}
\end{figure*}

%% file: sections/2_related_work.tex
\section{Related Work}
Spatial puzzle solving is closely related to Structure from Motion (SfM), pose estimation, arrangement learning, and more. The section discusses the related techniques.

%feature matching, pose estimation, arrangement learning
%involving feature matching, geometry inference

%\yasu{We do not say "extreme pose estimation" any more. Use "spatial puzzle" instead.}
%There are three approaches to the extreme pose estimation problems: feature matching, geometry inference, and arrangement learning.

\mypara{Feature matching} has been successful for the SfM problem~\cite{snavely2006photo,li20dualrc,Lin2016RepMatchRF}. The rise of deep neural networks enables more robust feature matching by learning~\cite{yi2018learning,sun2021loftr,arlin20superglue}. However, these techniques require visual overlaps. Our task has little to no visual overlaps between adjacent images.

\mypara{Geometry inference} estimates a relative pose between images or partial scans with minimal overlaps by registering inferred or hallucinated geometry. A popular approach learns the priors of room shapes and alternates pairwise alignment and scene completion~\cite{lin2019floorplan,Yang_2019_CVPR}. Yang~\etal ~\cite{yang2020extreme} combines global relative pose estimation and local pose refinement with panoramas. These techniques learn priors of a single room, while our approach learns the arrangements of multiple rooms on a house scale.

\mypara{Arrangement learning} is the current state-of-the-art for indoor room layout arrangement.
%, where 
%E-SfM.
%the main idea is to use doors/windows to align two images or reconstructions that might not have any overlaps.
%
An early work uses windows to align indoor and outdoor reconstructions~\cite{indoor-outdoor-3drec}.
% despite no learning on the arrangement itself.
Shabani~\etal~\cite{amin} 
% is the closest to ours, where they 
use doors to enumerate room arrangements and learn to score each candidate. Their approach is exponential in the number of rooms with many heuristics. Lambert~\etal~\cite{zillow} uses doors, windows, and openings to create room alignment hypotheses. They utilize depth maps to create top-down views and learn to verify the correctness, improving a run-time from exponential to polynomial.
Our end-to-end approach does not enumerate arrangement candidates and makes significant performance improvements. Lastly, 
%while not aligning doors nor windows,
an annotated site map and SfM reconstructions are aligned to solve a challenging structure from motion problem~\cite{martin20143d, hosseini2023floorplan}, which they coined as a ``3D jigsaw puzzle''.

\mypara{Puzzle Solving} has been an engaging research area for a long time \cite{freeman1964apictorial, radack1982jigsaw, markaki2022jigsaw}, ranging from pictorial puzzles with image information~\cite{shih2018divide, le2019jigsawnet, toler2010multi} to apictorial puzzles with only geometry information~\cite{goldberg2002global, harel2020lazy, hoff2014automatic, hoff2013extensions, harel2021crossing}. In previous studies, heuristic-based methods, such as edge and color matching~\cite{wolfson1988solving, nielsen2008solving, chung1998jigsaw}, have been predominantly used for solving both methods. 
%However, these heuristics are incapable of mastering high-level spatial reasoning.
%the reasoning involved in high-level arrangements.
%However, these heuristics are not capable of learning high-level arrangement reasoning.
%information.
%features.
%in pictorial puzzles.
%Additionally, they lack generalization and only work for a specific puzzle type.
Recently, deep learning-based methods seek to learn high-level pictorial features~\cite{noroozi2016unsupervised, li2021jigsawgan}. However, they suffer from poor geometric reasoning and are limited to simple square puzzles.
%, with a fixed classification output to predict the permutation of pieces.
Consequently, researchers combine learning-based models with heuristics to handle more complex puzzles~\cite{le2019jigsawnet}. Nonetheless, these methods still require explicit pairwise comparison of pieces and lack data-driven high-level reasoning . Our method overcomes these limitations through an innovative end-to-end use of Diffusion Models which are usually regarded as powerful generative models.
%being more robust to noise and erosion than heuristic methods, while also being applicable to various puzzle types.

\mypara{Diffusion models} (DMs) are emerging generative models, which slowly corrupt a sample by adding noise~\cite{denoising, beats,impdenoising, shabani2022housediffusion},
learn to invert the process, and generate a diverse set of samples from noise signals.
DMs have established SOTA performances in numerous tasks such as image colorization/inpainitng ~\cite{color, gild}, image to image translation~\cite{trans, trans2}, text to image~\cite{clip}, super-resolution~\cite{rombach2021highresolution, saharia2021, SRDiff}, image and semantic editing~\cite{avrahami2022blended_latent,meng2022sdedit}, 
and denoising~\cite{kawar2022denoising}. Recent works use DMs as representation learners for discriminative tasks such as image segmentation~\cite{baranchuk2021labelefficient, wolle}.
Diffusion inspired models have been used for human pose estimation~\cite{diffpose, difpose2}, and object placement~\cite{legonet}, however, both tasks do not require sophisticated shape comparisons. While in our case a more advanced and intricate approach is necessary to capture and exploit all details of the shapes.

%To our knowledge, the use of DMs for non-generative tasks has been limited. We are the first to apply DMs to the domain of spatial puzzle solving, and potentially, also the first in the broader areas of pose and arrangement estimation.
%However, to our knowledge, few works use DMs for non-generative tasks, where we are the first to use DMs for spatial puzzle solving and also the first even in broader areas of pose or arrangement estimation.

%pose estimation.

%% file: sections/3_dataset.tex
\section{Spatial Puzzle Solving Tasks}

Spatial puzzle solving involves arranging puzzle pieces, each with its geometry and optional features such as images or categories. In real scenarios, puzzle pieces are affected by erosion, duplication, or loss. This paper investigates three distinct puzzle-solving tasks.

$\bullet$ Cross-cut Jigsaw Puzzle (CJP) generates pieces by making random straight cuts through a larger polygonal shape~\cite{harel2021crossing}. 
Puzzle pieces are convex polygons, each with an arbitrary number of neighboring pieces. The sum of the two adjacent angles at any corner equals $180^\circ$. The application of the technique could be the restoration of shattered artwork.

$\bullet$ Voronoi Jigsaw Puzzle (VJP) generates pieces by 
%we start by
randomly sampling points within a predetermined bounding box and extracting the cells of the Voronoi diagram as the pieces.
%which are then used to generate the Voronoi diagram. The resulting regions are considered as the pieces of the puzzle. 
With the lack of the ``$180^\circ$ constraint'', VJP is significantly more challenging, where no effective solution has been presented in the literature to our knowledge. Voronoi diagrams play a role in biological systems like cell arrangements~\cite{cell}, potentially useful in studying natural phenomena.

$\bullet$ Room Layout Arrangement (RLA) determines the room arrangement and the corresponding floorplan, 
%from a set of indoor panoramas or room layouts
offering a key application in real estate.
%industry. Its objective 
% Compare to the previous puzzles,
Contrary to the prior tasks, a corner may align along the edge of another piece, expanding the solution space to be explored. Pieces come from the room layout estimation algorithms~\cite{amin,zillow} or interactive augmented reality apps used by consumers.

\mysubsubsection{Task input/output}
% \mysubsubsection{Input}
The input is a set of $N$ polygons (puzzle pieces in CJP/VJP and room layouts in RLA), each of which is a sequence of corner coordinates forming a 1D polygonal loop. In RLA, a door piece is also given as a line segment with two corners, and a room piece is associated with a room type as a 20D one-hot vector.
For a pictorial version of CJP, a piece
%, which is represented as a one-hot vector. We also consider pictorial version of VJP in which each piece
is associated with a 128D image feature vector 
%image feature of length 128 
obtained from a pretrained auto-encoder. 
%
%consisting of two door corners.
For simplicity, we mix room-corners and door-corners, and use $C^r_i$ to denote the 2D coordinate of the $i$th corner in the $r$th polygon.
$\mathcal{T}^r$ denotes the image feature or the room type vector.
Please refer to supplementary for details.
%
% \mysubsubsection{Output}
The output is the position of the piece/room center and the rotation around it (an angle between 0 and $2\pi$ for CJP/VJP and a 4-fold Manhattan rotation for RLA).
The center is the average of the corners.
%is the average of the corner coordinates.

\mysubsubsection{Metrics}
%\mysubsubsection{Metrics}
For CJP and VJP, we adopt metrics used in previous work~\cite{harel2020lazy}, namely Overlap, Precision, and Recall. The overlap score is the average IoU of pieces with the ground truth.
%respected to their new positions in the predicted arrangement, while 
Precision and Recall are on the connectivity of neighboring pieces.
%These metrics provide a comprehensive evaluation of the performance in capturing the visual characteristics of the objects.
%
For RLA, we consider two metrics. The first metric is the Mean Positional Error in pixels (MPE) over the rooms~\cite{amin}.~\footnote{Shabani et al.~\cite{amin} used a permissive metric (the availability of the ``correct" solution in the k results with a certain error-tolerance) as the task was challenging. We make great improvements and use a standard metric.}
%This work boosts the performance and uses a standard metric.}
The second metric evaluates the correctness of the room connectivity in the reconstruction. We borrow a Graph Edit Distance (GED) by Nauata \etal .~\cite{housegan}, which counts the number of user edits necessary to fix the connectivity graph. We declare that two rooms are connected if the door centers are within 5 pixels from the two rooms.
%
%While the metrics for the three tasks share some similarities, 
Note that we have designed task-specific metrics, respecting the methodologies of existing literature while providing a thorough evaluation of our performance.

%% file: sections/4_preliminaries.tex
%\section{Problem Definition}
% \yasu{Merge with the previous section}

% We borrow the problem formulation by Shabani \etal~\cite{amin} with minor adaptions to the input and the metrics:

% \mysubsubsection{Input} 
% Their original input was a set of panorama images, where a room layout estimation~\cite{HorizonNet} and an object detection network were used to convert the panoramas to top-down semantic images. In our datasets, room layouts do not come from panoramas but from an interactive AR application. Therefore, we define our input to be a set of $N$ room shapes, each of which is a sequence of room-corner coordinates forming a 1D polygonal loop in the top-down view. A door is a line segment consisting of two door corners. For simplicity, we mix room-corners and door-corners, and use $C^r_i$ to denote the 2D coordinate of the $i_{\mbox{th}}$ corner in the $r_{\mbox{th}}$ room.
% In order to remove the ground-truth position information, the mean room-corner coordinate is subtracted from all the corner coordinates for each room.
% A room is associated with a room type, which is represented as a one-hot vector $\mathcal{T}^r$.

% \mysubsubsection{Output}
% The output is the position of the room center 
% and the 4-fold rotation
% under the Manhattan assumption for each room.
% The rotation is around the room center, which is the average of the room corner coordinates.

%% file: sections/5_method.tex
%\section{Method}
\section{Spatial Puzzle Solving as Conditional Generation}
% \yasu{Regardless of the tasks, we want to use a consistent notation. Can we define one symbol for a condition information, and just define it for each task? Then, we can use that symbol to explain our system which works for any tasks. }

 \begin{figure*}
     \centering
\includegraphics[width=\textwidth]{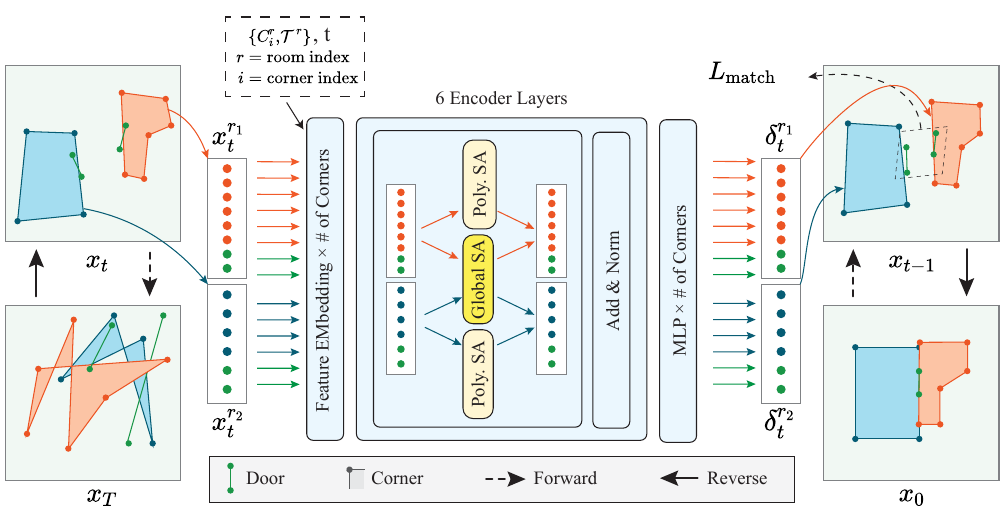}
     \caption{
     % A diffusion model architecture for Extreme Structure from Motion. 
Illustration of our diffusion model architecture employed for the RLA task.
     Given the arrangement estimation $x_t = \{x^r_{i,t}\}$, the reverse process infers the noise $\{\delta^r_{i,t}\}$ and recovers $x_{t-1}=\{x^r_{i,t-1}\}$, while injecting the original room shapes $\{C^r_i\}$ and types $\{\mathcal{T}^r\}$ as the condition.
     Each room corner holds the room position and rotation estimation.
     The reverse process starts from $x_T$ and denoises towards $x_0$.}     
     \label{fig:method}
 \end{figure*}
 
% Our idea is simple, using a Diffusion Model to ``conditionally generate'' a room-layout arrangement,
% where the input room layouts and types are the conditions.
Our idea is simple, using a Diffusion Model to ``conditionally generate'' the correct arrangement,
where the input is the center and rotation of room layouts/puzzle pieces and their types and shapes are the conditions.
This section explains the forward and the reverse processes.

%%%%%%%%%%%%%%%%%%%%%%%%%%%%%%%%%%%%%%%%%%%%%%%%%%%%%%%%
\subsection{Forward process}
\label{section:forward}
%%%%%%%%%%%%%%%%%%%%%%%%%%%%%%%%%%%%%%%%%%%%%%%%%%%%%%%

% The forward process adds a Gaussian noise to a room-layout arrangement.
% A compact representation would be per-room positions and rotations.
% Instead, we will use a redundant representation, where a room center position and a 4-fold rotation are stored at each room/door corner.
The forward process adds a Gaussian noise to an arrangement.
A compact representation would be per-polygon positions and rotations.
Instead, we will use a redundant representation, where a center position and a rotation are stored at each corner.

There are a few reasons. Our reverse process is based on a Transformer architecture where each position/rotation estimation becomes a node. Our approach 1) enriches the capacity of the arrangement representation (also an adaptive capacity, where a complex
polygon
% room
with more corners is given more capacity); 2) allows direct communications between 
% doors
corners/doors
for which we will have a 
% door-specific
specific
loss; and 3) makes it straightforward to combine with the condition (\ie, original corner coordinates and room types).
%%%%%%%%%%%%%%%%%%%%%%%%%
% \begin{figure}
%     \centering
%     \noindent\includegraphics[width=\linewidth]{figures/corner_2.pdf} 
%     \caption{Each room/door corner holds the room position $p^r_{i,t}$ and the rotation $o^r_{i,t}$ (not visualized here) estimation. At time $t=0$ before noise injection, they share the same ground-truth values per room. ${C^r_i}$ denotes the original room corner coordinates with respect to the room center.
%     $p^r_{i,t}+C^r_i$ indicates the estimated position of the $i$th corner of the $r$th room at time $t$.
%     }
%     \label{fig:input_output}
% \end{figure}
% \begin{wrapfigure}{l}{0.45\textwidth} % 'l' for left alignment, 'r' for right
%     \centering
%     \includegraphics[width=\linewidth]{figures/corner_2.pdf} 
%     \caption{Each room/door corner holds the room position $p^r_{i,t}$ and the rotation $o^r_{i,t}$ (not visualized here) estimation. At time $t=0$ before noise injection, they share the same ground-truth values per room. ${C^r_i}$ denotes the original room corner coordinates with respect to the room center. $p^r_{i,t}+C^r_i$ indicates the estimated position of the $i$th corner of the $r$th room at time $t$.}
%     \label{fig:input_output}
% \end{wrapfigure}
%%%%%%%%%%%%%%%%%%%%%%%%%

Concretely, we use $x^r_{i,t}$ to denote the position/rotation of the $r$th room/piece stored at the $i$th corner at time $t$ of the diffusion process, where $t$ varies from 0 to 1,000 in our experiments:
\begin{eqnarray}
x^r_{i,t} &=& \left(p^r_{i,t}, o^r_{i,t} \right).
\end{eqnarray}
% $p^r_{i,t}$ and $o^r_{i,t}$ denote the room-center position (a 2D vector) and the 4-fold rotation (a 4D one-hot vector).
$p^r_{i,t}$ and $o^r_{i,t}$ denote the polygon-center position (a 2D vector) and the rotation. We consider rotation as a 2D vector obtained from rotation matrix including cos($o^r_{i,t}$) and -sin($o^r_{i,t}$).
%Note that in RLA, $o^r_{i,t}$ has only 4 possible value assuming the Manhattan assumption.
The forward process adds a noise by sampling $\delta^r_{i,t}\in \mathcal{N}(\mathbf{0}, \mathbf{I})$
with a standard cosine noise scheduling with variance $(1 - \alpha_t)$~\cite{cosine}:
\begin{eqnarray}
x^r_{i,t} = \sqrt{ \bar \alpha_t} x^r_{i,0} + \sqrt{1 - \bar \alpha_t} \delta^r_{i,t} ,\quad \bar{\alpha}_t=\frac{f(t)}{f(0)}, \quad f(t)=\cos \left(\frac{t / T+0.008}{1+0.008} \cdot \frac{\pi}{2}\right)^2
\label{eq:forward}\end{eqnarray}

%%%%%%%%%%%%%%%%%%%%%%%%%%%%%%%%%%%%%%%%%%%%%%%%%%%%%%%%
\subsection{Reverse process}
%%%%%%%%%%%%%%%%%%%%%%%%%%%%%%%%%%%%%%%%%%%%%%%%%%%%%%%%
Figure~\ref{fig:method} illustrates our reverse process for room layout arrangement, which takes the arrangement $\{x^r_{i,t}\}$ at time $t$ and infers the noise $\tilde{\delta^r_{i,t}}$ under the condition of the original room shapes $\{C^r_i\}$ (\ie, a corner position with respect to the room center) and the room types as a one-hot vector $\{\mathcal{T}^r\}$.

\mysubsubsection{Feature embedding}
The reverse process is based on a Transformer architecture where every corner is a node. We initialize its feature embedding as
\begin{eqnarray}
\hat{x}^r_{i,t} \leftarrow 
 \mbox{Lin}(x^r_{i,t}) + \mbox{Lin}([C^r_i, r ,i]) + \mbox{MLP}(t) + \mbox{Lin}( \mathcal{T}^r). \label{eq:reverse}
\end{eqnarray}
The first term uses a linear layer to convert a 4D vector (\ie, 2 for the room center coordinate and 2 for the rotation vector) to a 256D embedding vector. The second term also uses a linear layer to convert a 
% 28d 
condition vector 
(\ie, 2 for the corner coordinate respecting to the center of the polygon in addition to 32 for polygon index which shows which polygon each corner belongs $r$ and 32 to show the corner index in each polygon $i$),  The third term uses a 2-layer MLP to convert a time step $t$ to a 256D vector. The fourth term is extra information such as room/door type (20 types)  for RLA which we extract it by applying a linear layer on 20D one-hot vector or 128D for image features in pictorial CJP) 

% The fourth term is a standard frequency position encoding~\cite{vaswani2017attention} of a corner index $i^\prime$. Note that this corner index is across all polygons instead of one polygons, where polygons orders are random. 
% An alternative way is to use linear layers to embed a polygons index and a corner index. Our index mixes polygons and does not indicate which corners belong to the same polygons, but is free from any network parameters to learn and works well in practice. Our Transformer architecture injects corner-to-polygons association information by structured self-attentions next.

\mysubsubsection{Attention modules}
Feature embeddings $\{\hat{x}^r_{i,t}\}$ go through six blocks of self-attention modules that have two different attention mechanisms:
Polygon Self Attention (P-SA) and Global Self Attention (G-SA).
P-SA limits pairwise interactions between corners in each polygon.
P-SA is akin to a sparse self attention family~\cite{guo2019star, child2019generating, li2019enhancing}, which helps to generate consistent positions and rotations at different corners of a polygon.
G-SA is a standard self-attention between all corners of a puzzle or a house.
After the attention blocks, a linear layer converts 256D embedding back to a 4D representation $\tilde{\delta}^r_{i,t}$, which is used for the following denoising formula~\cite{denoising}:
\begin{eqnarray}
x^r_{i, t-1}=\frac{1}{\sqrt{\alpha_t}}\left(x^r_{i,t}-\frac{1-\alpha_t}{\sqrt{1-\bar{\alpha}_t}} 
\tilde{\delta}^r_{i,t}
\right)+\sqrt{1-\alpha_t} z.
\end{eqnarray}
$z  \sim \mathcal{N}({0}, {I}) $ for $t>1$ and otherwise 0. For the final result at time $t=0$, we take the average polygon center position and the polygon rotation.
% (in fact, the majority vote after picking the Manhattan-rotation with the highest score at each corner).

\mysubsubsection{Loss functions}
There are two loss functions. We first follow~\cite{denoising, beats} and use a standard noise regression loss on $\delta$:
\begin{equation}
    L_{\text{simple }}=E_{t, x^r_{i,0}, \delta^r_{i,t}}\left[\left\|\delta^r_{i,t}-\tilde{\delta}^r_{i,t}\right\|^2\right].
  \label{eq1}
\end{equation}

% The second loss is on door corners, each of which is shared by two rooms.
% Let us use $r_1$ and $r_2$ to denote its room indexes. 
% $i_1$ and $i_2$ denote the corresponding door-corners indexes in $r_1$ and $r_2$.
% The corresponding door corners must be at the same position and the loss is the L2 distance of their positions:
% \begin{eqnarray}
%   L_{\text{match}}&=&E_{t, x^r_{i,0}, \delta^r_{i,t}}\left[\left\| \hat{C}^{r_1}_{i_1} - \hat{C}^{r_2}_{i_2} \right\|^2 \right],
%     \label{eq2} \\
% \hat{C}^r_{i}&=&\tilde{p}^r_{i,0} + R_{\tilde{o}^r_{i,0}}  C^r_i, \\
% (\tilde{p}^r_{i,0}, \tilde{o}^r_{i,0}) &=& \tilde{x}^r_{i,0} =\left(x^r_{i,t}-\sqrt{1-{\bar \alpha_t }} \tilde{\delta}^r_{i,t} \right) / \sqrt{{\bar \alpha_t}}.
% \label{eq:xtox_0}
% \end{eqnarray}
% $R_{\tilde{o}^r_{i,0}}$ denotes a Manhattan rotation matrix corresponding to the largest entry in $\tilde{o}^r_{i,0}$. Note that obtaining the rotation matrix
% $R$ is not differentiable and the gradients are not propagated. The rotation branch is only trained with the first noise regression loss.
% The total loss is defined as
% $L_{\text{total}}= L_{\text{simple}} + 0.01 \cdot L_{\text{match}}$.

To enhance the quality of supervision, we propose a ``matching'' loss, specifically aimed at the vertices where incident edges meet. These vertices are shared by two polygons. We denote the indices of these two polygons as $r_1$ and $r_2$, then the corresponding vertex indices within $r_1$ and $r_2$ as $i_1$ and $i_2$, respectively. The corresponding vertices must be at the same position, and as such, we calculate the loss as the Euclidean distance between their coordinates:
\begin{eqnarray}
  L_{\text{match}}&=&E_{t, x^r_{i,0}, \delta^r_{i,t}}\left[\left\| \hat{C}^{r_1}_{i_1} - \hat{C}^{r_2}_{i_2} \right\|^2 \right],
    \label{eq2} \\
\hat{C}^r_{i}&=&\tilde{p}^r_{i,0} + R_{\tilde{o}^r_{i,0}}  C^r_i, \\
(\tilde{p}^r_{i,0}, \tilde{o}^r_{i,0}) &=& \tilde{x}^r_{i,0} =\left(x^r_{i,t}-\sqrt{1-{\bar \alpha_t }} \tilde{\delta}^r_{i,t} \right) / \sqrt{{\bar \alpha_t}}.
\label{eq:xtox_0}
\end{eqnarray}
$R_{\tilde{o}^r_{i,0}}$ denotes the rotation matrix corresponding to $\tilde{o}^r_{i,0}$. For the room layout arrangement task, we add the loss only to doors without walls, which yields superior results in our experiments.
%door-based supervision, as opposed to relying on shared walls (i.e polygon edges), leads to better performance. 
The total loss is defined as the sum of the above loss functions, $L_{\text{total}}= L_{\text{simple}} + L_{\text{match}}$. In practice, we found that adding $L_{\text{match}}$ only for $t<500$ results in better performance. 

%% file: sections/7_experiments.tex
%%%%%%%%%%%%%%%%%%%%%%%%%%%%%%%%%%%
% \begin{figure*}[tb]
% \begin{minipage}[t]{.5\textwidth}
%     \centering
%  \noindent\includegraphics[width=\textwidth]{figures/main_res_new.pdf} 
%     \caption{Qualitative evaluations of our approach against the three competing methods. The top two rows are from Small JigsawPlan. The bottom row is from Small RPLAN. The GT rotations are given for all the cases to enable comparisons with all the methods.}
%     \label{fig:main}
% \end{minipage}
% \end{figure*}
% %%%%%%%%%%%%%%%%%%%%%%%%%%%%%%%%%%%
% \begin{figure}[tb]
% \begin{minipage}[t]{.5\textwidth}
%     \centering
%  \noindent\includegraphics[width=0.99\linewidth]{figures/shab_ours.pdf} 
%     \caption{Qualitative evaluations when the GT rotations are not given. The top row is from Small JigsawPlan. The bottom row is from Small RPLAN.    }
%     \label{fig:ours-vs-amin}
% \end{minipage}
% \end{figure}
%%%%%%%%%%%%%%%%%%%%%%%%%%%%%%%%%%%
\section{Experimental Results and Discussions}
We have implemented the system with PyTorch~\cite{paszke2019pytorch}, using a workstation with a 3.70GHz Intel i9-10900X CPU (20 cores) and two NVIDIA RTX A6000 GPUs. We use the AdamW~\cite{loshchilov2017decoupled, kingma2014adam} optimizer with $\beta_1=0.9$, $\beta_2=0.999$, weight decay equal to 0.05, and the batch size of 512. The learning rate is initialized to 0.0005.
% The pose estimation process is stochastic and we run our system 5 times and report the mean and the standard deviation.
It takes roughly 24 hours to train a model and 3 seconds to estimate the arrangement for one sample.

\subsection{Preprocessing and Datasets}
%\mysubsubsection{Coordinate scaling}
% Our benchmarks for RLA are constructed from the ground-truth floorplans whose scales were normalized, so that entire floorplans fit inside a $256\times 256$ square. This could allow cheating because the longer extent of arranged floorplans is always $255$ pixels. Therefore, at test time, we scale the room shapes by a random scaling factor in the range of [0.8, 1.0] for each house.
We have carefully pre-processed and prepared datasets for fair evaluation, whose details are referred to supplementary. We here provide summary points for the three tasks (See Fig.~\ref{fig:dataset}).

%For each task, we carefully preprocess and prepare datasets for fair evaluation, whose details, including noise levels and pictorial feature extraction, are referred to supplementary.
%in addition to other information such as noise levels and pictorial feature extraction.

\mypara{Cross-cut Jigsaw Puzzle (CJP)}:
We have used code provided by Harel \etal~\cite{harel2021crossing} to generate 100k/1k puzzles for training/testing. 
The code generates a convex polygon and cuts it by 3 to 5 lines. For the pictorial version, we have used images from COCO 2017 dataset~\cite{coco2017}, that is, randomly selecting a training/testing image of the dataset for each training/testing CJP sample. Following~\cite{harel2021crossing}, to simulate real-world scenario,  we perturb corner coordinates with noise with three different levels to create datasets under three different noise levels with thresholds equal to 0, 2, and 4. 

%sed coco 2017 dataset/ we used coco 2017 train images for our train ( randomly selecting 100 K images from them )and coco2017 test (randomly selecting 1K images from it).

%we pick an image from ImageNet database randomly~\cite{imagenet}.
%For CJP, we generate 100k puzzles for training and 1k puzzles for the test by randomly cutting a convex polygon using 3 to 5 straight lines~\cite{harel2021crossing}.

\mypara{Voronoi Jigsaw Puzzle (VJP)}: We have generated 200k/1k puzzles for training/testing by randomly sampling 3 to 15 points inside a random rectangle and extracting their Voronoi cells as the pieces. We have created datasets with three different levels of noise. For each corner, we added a random number from a Gaussian distribution with $\sigma^2$ equal to 0, 2, and 4 for noise levels 0, 1, and 2, respectively.
%considering the pieces in their corresponding Voronoi diagram.
%

\mypara{Room Layout Arrangement (RLA)}:
%Unlike previous tasks that involve generating synthetic and easily accessible samples, obtaining a real-world annotated dataset for RLA is costly due to concerns regarding privacy and the extensive efforts required to gather such data.
MagicPlan (https://www.magicplan.app), a mobile software company for real estate and construction, agrees to share production data with us and the research community, where the paper introduces the MagicPlan dataset,
%In this regard, we introduce a new dataset, MagicPlan,
containing room shapes and their ground-truth arrangement for 98,780 single-story houses/apartments. We split the data into 93,780/5,000 training/testing samples.
MagicPlan software reconstructs room shapes one by one by asking users to click room corners through an augmented reality app.
Room shapes are Manhattan-rectified as an enforcement of the app, then manually arranged to form a floorplan, which we seek to automate.
%
%An Augmented Reality (AR) application creates room layouts by asking users to click room corners. 
%TNote that the layouts are Manhattan-rectified as an enforcement of the app.
Each room is associated with a room type. The number of rooms (resp. corners) in a house ranges from 3 to 10 (resp. 12 to 182).
%, and the minimum and the maximum numbers of corners in a house are 12 and 182.
% Our second RLA benchmark is based on a floorplan dataset, RPLAN, to demonstrate robustness across different datasets. 
We also use RPLAN~\cite{wu2019data} for evaluation, containing 60k floorplans. We divide into 55k/5k training/testing, where the number of rooms (resp. corners) in a house ranges from 3 to 8 (resp. 14 to 98).
%RPLAN contains 60K vector floorplans made by architects. Among which, we consider 55K for the training and 5K for the test. The number of rooms in a house ranges from 3 to 8. The number of corners ranges from 14 to 98. 
%See the supplementary for more details on the datasets. 
Note that RICOH dataset~\cite{saharia2021} and ZIND dataset~\cite{zind_sfm} are too small for network training and are not used in our experiments.

\subsection{Competing methods} \label{sec:competing_methods}
% \yasu{May need to explain per task?}
% Their solution exploits specific geometric properties: plausible matings require identical edge lengths and vertices emerging only from two crossing cuts. This implies that the pairs of adjacent angles from neighboring pieces must sum to 180 degrees.

Harel \etal~\cite{harel2021crossing} and Shabani \etal~\cite{amin} are state-of-the-art methods for CJP and RLA, respectively. We have used their public implementations to compare in the corresponding tasks. Note that Harel \etal is not applicable to VJP or RLA, where neighboring angles may not add to $180^\circ$ or corners may not meet. Shabani~\etal is not applicable to CJP or BJP due to its poor scalability (i.e., exponential in the number of pieces). 
We have also prepared a third method based on the transformer network to be compared for RLA. The following provides more information, while the full details are in supplementary.
%of each method.

%We consider two main baselines to compare our method: Harel \etal~\cite{harel2021crossing} for CJP and Shabani \etal~\cite{amin} for RLA. It is worth noting that while the method proposed by \cite{amin} can theoretically be used for CJP and VJP as well, it is infeasible due to the large number of pieces involved in these tasks. Furthermore, we compare our method with TransVector, another baseline, on RLA to further demonstrate the effectiveness of the DM formulation.

\mypara{Harel \etal}~\cite{harel2021crossing} proposed a two-step algorithm for CJP: Enumerating pairs of compatible pieces by heuristics based on edge lengths or corner angles, then globally solving for the whole arrangement by a spring system.

%%. The first step involves
%Finding the candidate mates by utilizing the specific geometry properties of Cross-cut puzzles. In the second step, a spring system-based approach is employed to determine the correct position and rotation of each piece.

\mypara{Shabani \etal}~\cite{amin} enumerates arrangement candidates by heuristics and learns to regress the realism of an arrangement candidate by deep neural networks.
%is the existing state-of-the-art for RLA, which enumerates arrangement candidates and learns to classify a layout candidate to be realistic or not. We use their public implementation\footnote{\href{https://github.com/aminshabani/extreme-indoor-sfm}{https://github.com/aminshabani/extreme-indoor-sfm}}, where RPLAN and MagicPlan datasets have a different number of room/door types,
Since the number of room/door types differs in our datasets, we made minor modifications to the data loader and the network architecture.
%to adjust the feature dimension.

\mypara{TransVector} is a baseline Transformer network with a vector representation that directly estimates the pose parameters instead of iterative denoising. TransVector shares the same architecture as our denoising network at the core with the following changes:
%
%the second baseline for RLA, which uses our transformer network module at the core without the Diffusion.
% The baseline
%without iterations, and we make the following changes to our architecture:
1) Remove time-dependent features ($x^r_{i,t}$ and $t$) from the embedding (\ref{eq:reverse}); 2) Change the supervision $\delta^r_{i,t}$ from the noise to the position/rotation parameters; and We have also compared with the Transformer network with a raster representation, which is presented in the supplementary.

%. We refer details of the baseline systems to the supplementary.

%\mypara{Transformer with a raster representation} (TransRaster) uses the raster images to represent the input room layouts/types and the output room positions. Note that this baseline does not handle rotations as explained below.
%An input room layout is represented as a 25-channel $256 \times 256$ semantic segmentation image, where there are 25 room/door types. The room center is aligned with the center of an image.
%An output room position is represented as a $256 \times 256$ room occupancy image, which is ideally a translated version of the input room segmentation image at the correct room location.
%Given an output room occupancy image, we perform an exhaustive search over the possible room translations and find one with the most overlap between the occupancy image and the translated room segmentation image.~\footnote{We could expand the search space with possible room rotations, but rooms are often symmetric. To be simple, we use this baseline only for experiments when ground-truth rotations are given.}
%
%We use VisionTransformer~\cite{vit} with a CNN decoder that takes a set of input room segmentation images and produces a set of room occupancy images.

%%%%%%%%%%%%%%%%%%%%%%%%%%%%%%%%%%%
\input{tables/main_quantitative_fp}
\begin{figure}[tb]
% \begin{minipage}[t]{.59\textwidth}
%     \centering
%     \includegraphics[width=\linewidth]{figures/main_res_new.pdf} 
%     \caption{Qualitative evaluations of our approach against the three competing methods. The top two rows are from Small JigsawPlan. The bottom row is from Small RPLAN. The GT rotations are given for all the cases to enable comparisons with all the methods.}
%     \label{fig:main}
% \end{minipage}\hspace{0.2cm} % Add horizontal space here
% \begin{minipage}[t]{.37\textwidth}
\begin{minipage}[t]{.545\textwidth}
    \centering
    \includegraphics[width=\linewidth]{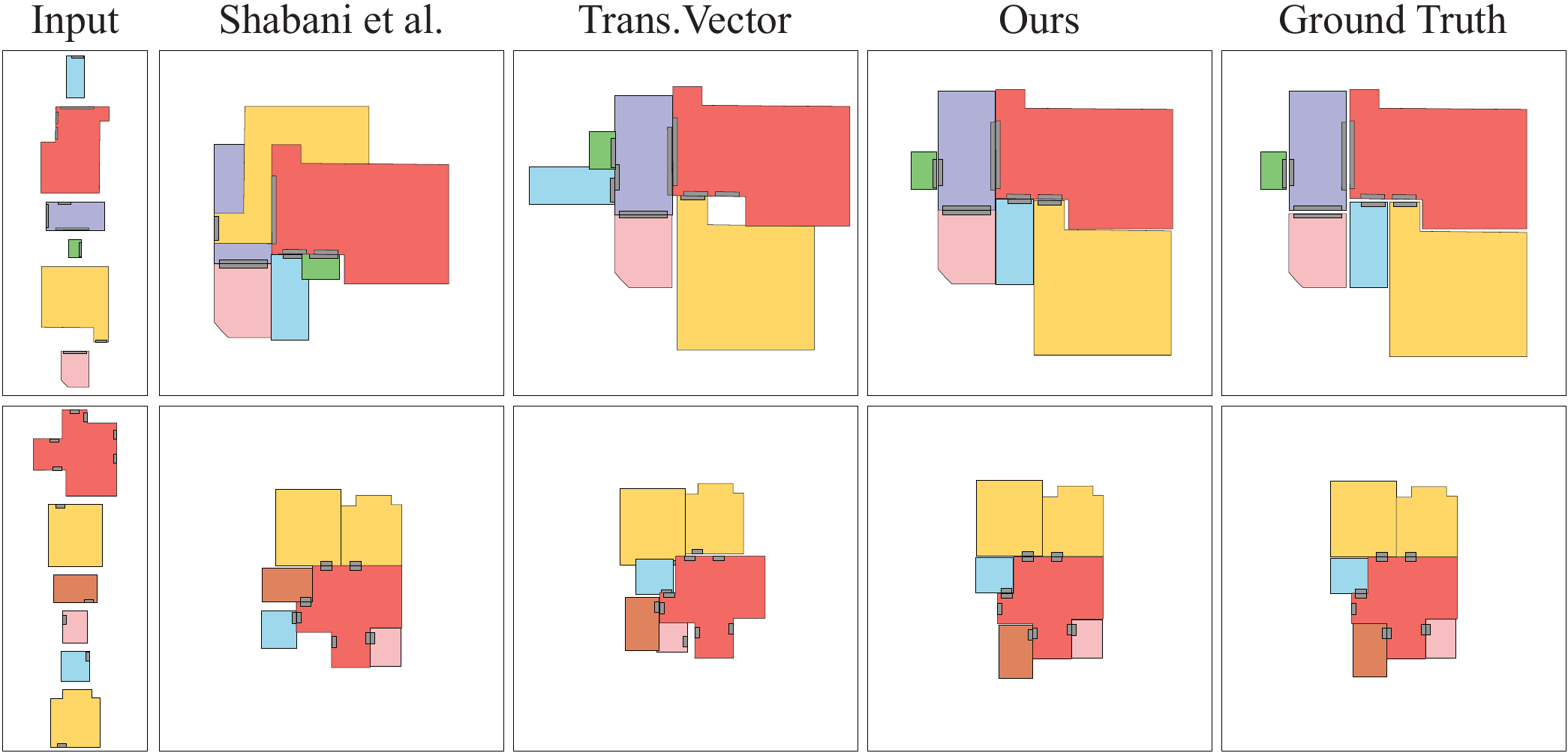} 
    \caption{RLA arrangement results 
    %of our method with the baselines for Room Layout Arrangement 
    with Small MagicPlan (top) and Small RPLAN (bottom).}
    % \caption{ RLA qualitative evaluations. The top row is from Small JigsawPlan. The bottom row is from Small RPLAN.}
    \label{fig:ours-vs-amin}
\end{minipage}
\hspace{0.2cm}
\begin{minipage}[t]{.42\textwidth}
   \centering
    \includegraphics[width=\linewidth]{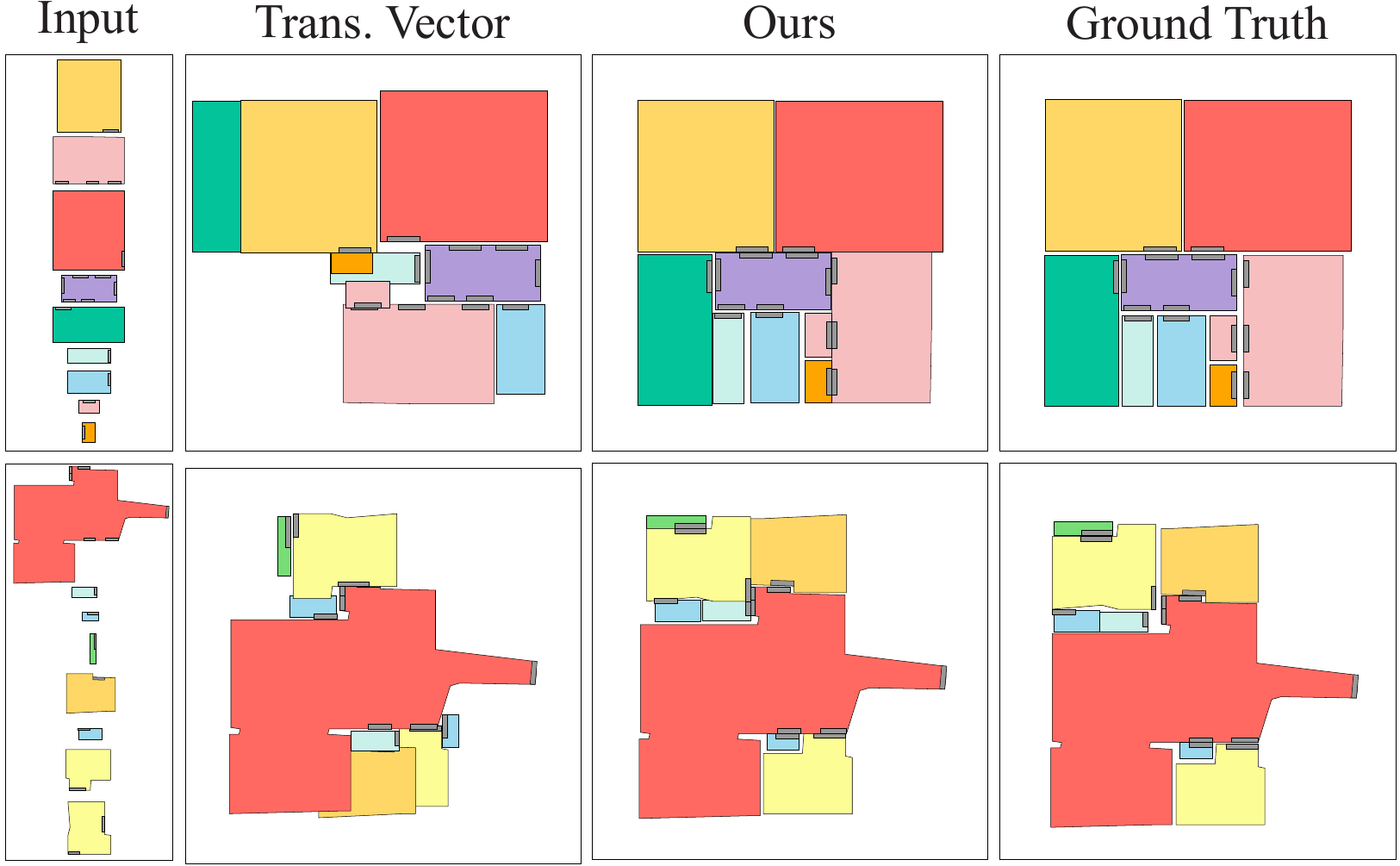} 
    % \caption{RLA qualitative evaluations. The top row is from Small JigsawPlan. The bottom row is from Small RPLAN.}
    \caption{RLA arrangement results with Full MagicPlan dataset.
    % Shabani \etal is infeasible here, duo to the exponential increase in number of possible arrangements.
    }
    \label{fig:s-f}
\end{minipage}
\end{figure}

\subsection{Main results}
%In this section, we first explain the results of our method on RLA, followed by the results on CJP and VJP.
Table~\ref{table:main} compares the proposed approach against Shabani \etal~\cite{amin} and TransVector.
%compares the MPE and the GED for the three methods. 
% TransRaster does not handle rotations, and the first group of four rows shows a case when room layouts are fixed to the ground-truth orientations. We modify the architecture and remove features and the loss function related to the rotation.
%bout the rotations
% The second group of three rows shows the full pose estimation results where inferring both the positions and the rotations,  room rotations are randomly initialized for each room.
% Room rotations are randomly initialized for each room.
%
Shabani \etal runs exponentially in the number of pieces (i.e., rooms), taking hours or even days to process a single house with seven or more rooms. Therefore, we collect small houses (\ie, at most six rooms) to create ``Small RPLAN'' and ``Small MagicPlan'' datasets for its evaluations.
%The left and the right of the table show results for RPLAN and MagicPlan datasets, respectively. Since the run-time of Shabani \etal~\cite{amin} is exponential in the number of rooms, it takes hours or even days to process a single house with seven or more rooms. Therefore, we extract small houses 
%
For each experimental setting (\eg, Small MagicPlan), we train a network for each method.
% Shabani \etal is evaluated only with Small RPLAN and Small MagicPlan.

Our method outperforms all the others in all the settings and metrics, except the GED metric for Small MagicPlan, where the existing state-of-the-art Shabani \etal achieves a slightly better score.
Their method enumerates all possible candidate arrangements by matching doors and evaluates the realism of each arrangement one by one. Their run-time is exponential in the number of rooms, and the system involves many heuristics.
TransVector is end-to-end and achieves comparable performance with Shabani \etal; in fact, a much better MPE score for Small RPLAN. However, our system performs much better in every metric, demonstrating the power of Diffusion Models even for non-generative tasks, in our case, room layout arrangement.
%Figure~\ref{fig:main} shows qualitative comparisons of all methods for the ``Small'' datasets with the given ground-truth rotations. Note that Shabani \etal cannot handle Full datasets.
%, and TransRaster requires ground-truth rotations. 
%This is the easiest setting (\ie, small houses with ground-truth rotations) where the competing methods produce reasonable results. Our approach produces the best, especially the middle example, where the arrangement of the three rooms on the right (blue, purple, and ping) is challenging.
%
Figure~\ref{fig:ours-vs-amin} compares Shabani \etal, Transvector and ours again on the small datasets. Figure~\ref{fig:s-f} shows our results for the most challenging setting (\ie, Full MagicPlan dataset) comparing to Transformer with vector representation.

\begin{figure*}[!t]
    \centering
 \noindent\includegraphics[width=0.99\textwidth]{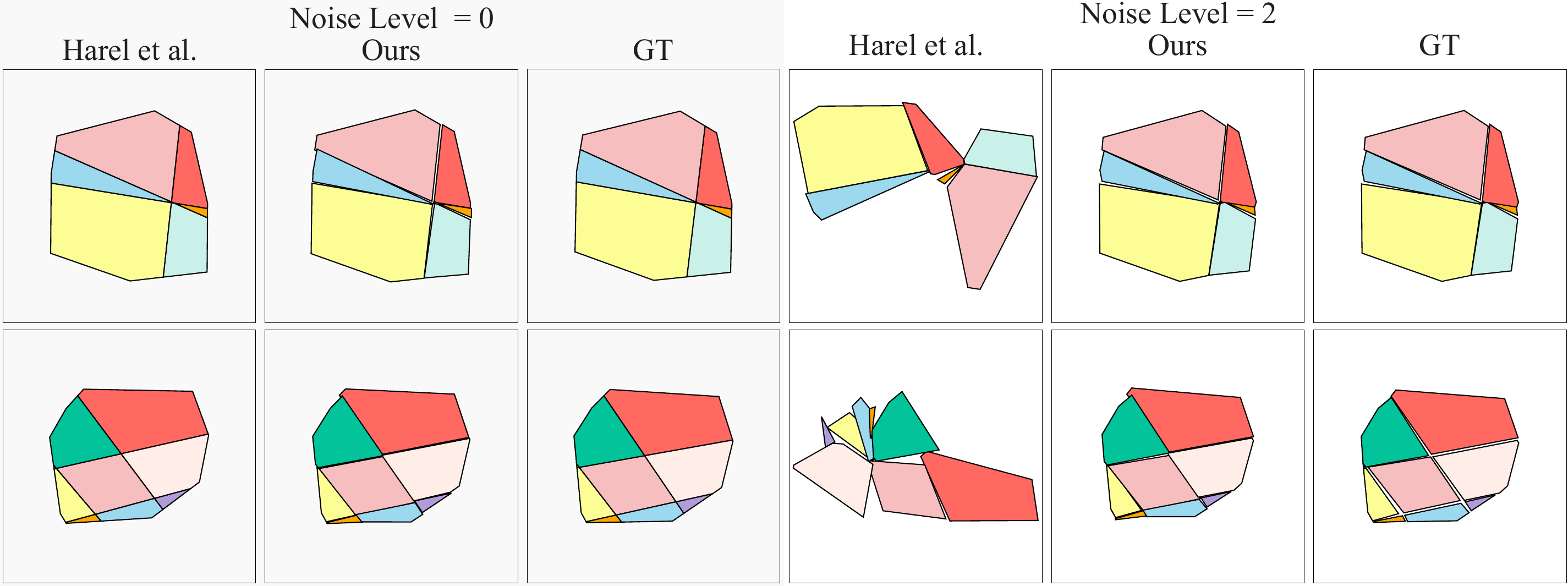} 
    % \caption{Apictorial CJP qualitative evaluations. Three left columns are samples without noise, and the three right columns are noisy samples.}
    \caption{Apictorial CJP arrangement results. Both methods work well in a noise-free case (left), but only our method maintains the performance with the presence of noise (right).}
    \label{fig:crosscutap}
\end{figure*}

\begin{figure}[t]
\begin{minipage}[t]{.44\textwidth}
    \centering
    \includegraphics[width=\linewidth]{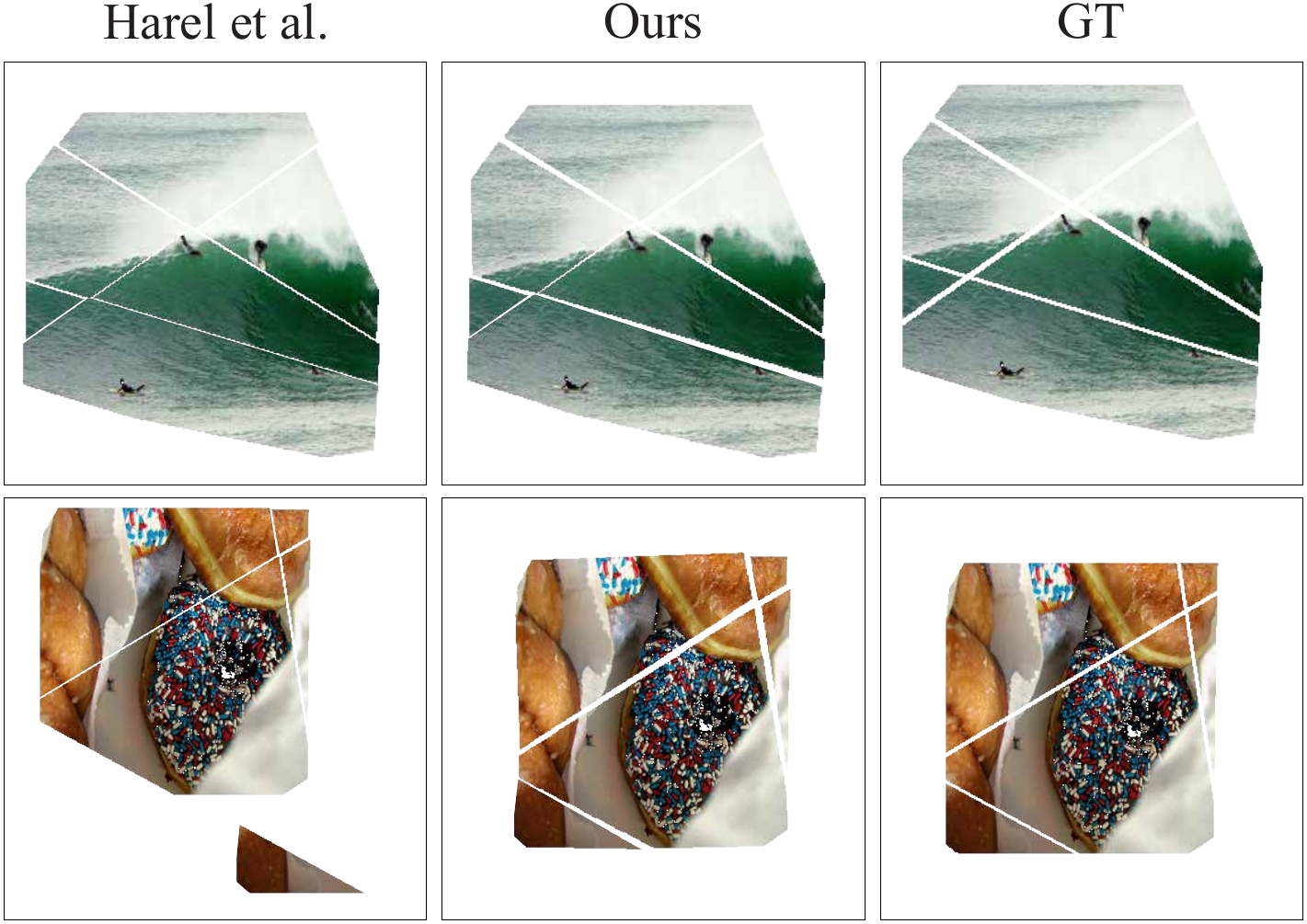} 
    \caption{Pictorial CJP arrangement results.}
    \label{fig:cross_pic}
\end{minipage}
\hspace{0cm} % Add horizontal space here
\begin{minipage}[t]{.54\textwidth}
    \centering
    \includegraphics[width=\textwidth]{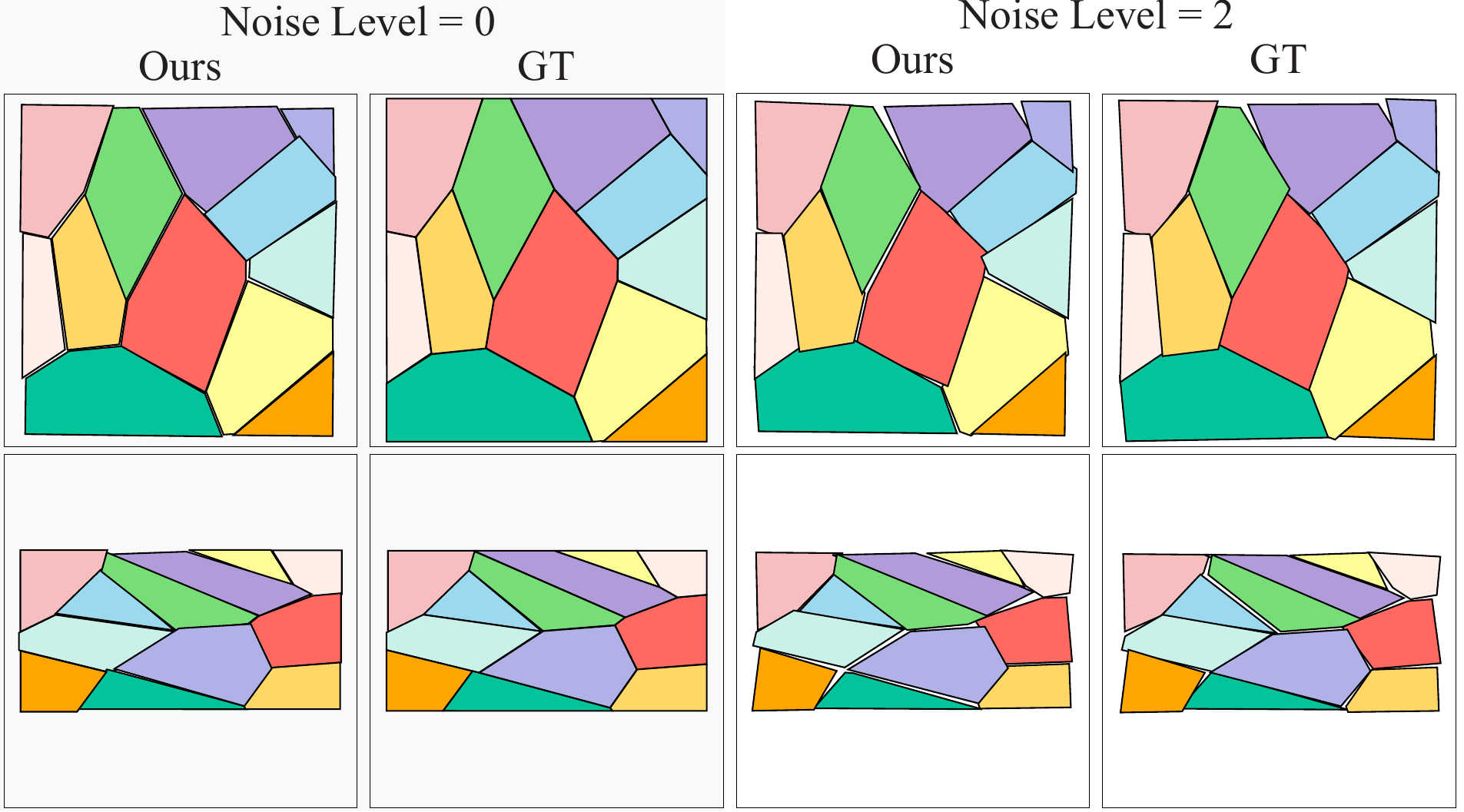} 
    % \caption{VJP qualitative evaluations. Two left columns  are samples without noise, and two right columns are noisy samples.}
    \caption{VJP arrangement results.}
    \label{fig:vor_pic}
\end{minipage}
\end{figure}

Table~\ref{table:mainp} presents a comparative analysis for CJP and VJP at three different noise levels. We have trained our model only for a noise-free case (i.e., 0 noise level) and used for the other two levels. Our approach consistently outperforms the current state-of-the-art. The performance gap is significant with the presence of noise, because Harel \etal relies on critical assumptions that matching edges have the same distance and neighboring angles sum to $180^\circ$. They have some tolerance to cope with noisy inputs but do not do well against our learning-based approach. Figs.~\ref{fig:crosscutap}, ~\ref{fig:cross_pic}, and ~\ref{fig:vor_pic} show sample arrangement results for apictorial CJP, pictorial CJP, and VJP, respectively, supporting the above quantitative results.
%The table displays our results in solving both noise-free and noisy puzzles, compared to the results by Harel \etal~\cite{harel2021crossing}. Note that we do not retrain our model on noisy samples and we use the trained model on the clean samples.
%showing that our method is significantly more robust to erosion compared to the baseline duo to our data-driven approach compared to heuristics in \cite{harel2021crossing}.
%Additionally, the table demonstrates the effectiveness of our method in solving Voronoi puzzles, where Harel \etal~\cite{harel2021crossing} is not applicable duo to its unique setting.
% Additionally, it demonstrates the effectiveness of our method in solving Voronoi puzzles, which is a unique setting where Harel et al.~\cite{harel2021crossing} were unable to achieve success. 
% In Figs.\ref{fig:crosscutap} and~\ref{fig:cross_pic}, we compare our performance on apictorial and pictorial cross-cut puzzles to that of other baseline methods. Fig.~\ref{fig:vor_pic} shows our method's robustness to different noise settings while solving Voronoi puzzles. We also have provided ablation studies on duplicate and missing pieces effect on puzzle solving in supplements.
Note that the table shows the numbers for apictorial CJP. Due to the time complexity of Harel \etal, we have only calculated the metrics on 20 samples for pictorial CJP, showing the same trend in results. The full details are referred to the supplementary, which also contains experiments on duplicate or missing pieces, where our method is surprisingly robust.
%
%Quantitatively, we evaluate our method on 20 pictorial puzzles, which is limited duo to the time complexity of ~\cite{harel2021crossing},  0.95 overlap score compared to 0.94 of Harel \etal~\cite{harel2021crossing}.
%We have also conducted experiments on duplicate or missing pieces
%We also have provided experiments on larger number of picotiral puzzles for our method and the ablation studies on duplicate and missing pieces effect on puzzle solving in supplements.
%where our method demonstrates robustness against such perturbations
%
%, we compare our performance on apictorial Cross-cut and Voronoi puzzles to that of other baseline methods. Fig.~\ref{fig:cross_pic} shows the qualitative results on pictorial VJP. 

%Our failures are often attributed to 1) Rare building architecture (top-left of Fig.~\ref{fig:s-f}) and 2) Inherent ambiguity (top-left of Fig.~\ref{fig:s-f}), whose tasks are challenging even for humans.

\subsection{Ablation studies}
We choose the room layout arrangement task for further ablation studies on our method.
%Room Layout Arrangement as the more challenging task to further demonstrate the effectiveness of our contributions.
%
Table~\ref{table:main_ab} shows the contributions of our two attention mechanisms (P-SA, G-SA) and the door matching loss.
G-SA provides communications between every pair of corners in a house, and its removal has the most impact on the performance.
Removing P-SA or the matching loss also leads to a significant drop in both MPE and GED metrics.
Our method with all the components achieves the highest performance.

\input{tables/puzzle_main}

\input{tables/ablation_attention}

\input{tables/ablation_roomtype}
Our layout arrangement uses a redundant representation (i.e., all the corners store the position and the orientation estimates of a piece), enriching the capacity and enabling direct communications between room/door corners (Sect.~\ref{section:forward}).
%
%Figure~\ref{fig:ensemble} shows the raw estimated position information at each room/door corner before the room-wise averaging. Since the ground-truth has the same pose parameters for all corners in a room/door, the network learns to produce consistent parameters.
%
To assess the effects of this redundancy, we have created a variant of our system with a compact representation, that is, each room/door has only a single node estimating a single copy of the room position and the orientation. We aggregate corner coordinates into a single embedding vector and pass as a condition (See supplementary for the details). The MPE/GED metrics for Full MagicPlan change from ($41.23/3.16$) to ($51.62/5.52$), a significant performance drop showing the importance of our redundant representation.

A popular approach in the RLA literature 
% indoor Extreme Structure-from-Motion 
is to align door detections/annotations to enumerate arrangement candidates~\cite{zillow,amin}. 
%Such approaches are susceptible to errors in the door information.
Our approach directly learns to infer the arrangement without relying on such ``hard'' heuristics.
%
%trains a network to infer the room arrangements instead of using heuristics to enumerate them, and is robust to input errors such as missing doors or incorrect room type annotations, which often happen in real world applications.
%
To further demonstrate the power of our approach,
%robustness of our approach, 
we remove or alter the room type and the door information during training and testing, where the current state-of-the-art methods~\cite{zillow,amin} completely fail. Table~\ref{table:door} shows some performance drop, but the effects are marginal. Our numbers are still much better than TransVector with the full information (MPE=$53.11$ and GED=$6.41$ in Table~\ref{table:main}), the only competing method capable of handling this most challenging setting.
To evaluate the capacity of our network to handle overlaps, we conducted an experiment. In this experiment, we augmented our train/test dataset by randomly selecting up to two rooms per house. For each selected room, we employed one of the following strategies: 1) Duplicating the room with the ground truth (GT) room types. 2) Duplicating the room with random room types. 3) Enlarging the room by 20$\%$ to create partial overlaps. The dataset used for this experiment is Full MagicPlan dataset. Our evaluation metrics, MPE/GED, resulted in 48.3/3.9, respectively. Please also see the supplementary document and the video for more results, more visualizations, and animations of the denoising process.
%more results and visualizations.

%Lastly, Figure~\ref{fig:diffrent_run} shows five pose estimation results by our system while varying the initial noise $x_T$. While there are minor differences, the overall room arrangements are similar and close to the ground-truth, indicating that the Diffusion model is capable of producing consistent results given enough constraints as a pose estimation system, as opposed to a generative model whose original goal is to create a diverse set of answers. 

%% file: tables/main_quantitative_fp.tex
\begin{table*}[!tb]
\centering
\caption{RLA quantitative results with two metrics: 
Positional Error (MPE) and Graph Editing Distance (GED).
%A group of four rows in the middle show a case where the ground-truth rotations are given. A group of three rows at the bottom show a case where both the positions and the rotations are estimated, where TransRaster baseline cannot handle rotations and is not evaluated.
Small RPLAN (resp. Small MagicPlan) is a subset of the corresponding full dataset, consisting of houses with at most 6 rooms. The small datasets are created for Shabani \etal~\cite{amin}, which is not scalable to many rooms.  %We run our system 5 times and
% SD for our method was less than 1.2 and 0.5 for MPE/GED respectively. }
}
\vspace{2mm}
\scalebox{0.85}{\begin{tabular}{l|c|c|c|c!{\vrule width 1.5pt}c|c|c|c!}
  \toprule
     \multicolumn{1}{l}{Dataset} & \multicolumn{2}{c}{Small RPLAN} & \multicolumn{2}{c}{Full RPLAN} & \multicolumn{2}{c}{Small MagicPlan }& \multicolumn{2}{c}{Full MagicPlan}\\
       \cmidrule(lr){1-1}
        \cmidrule(lr){2-3} \cmidrule(lr){4-5} \cmidrule(lr){6-7} \cmidrule(lr){8-9} 
	\multicolumn{1}{l}{Metric} & MPE $(\downarrow)$ & GED $(\downarrow)$ & MPE $(\downarrow)$ & GED $(\downarrow)$ & MPE $(\downarrow)$ & GED $(\downarrow)$ & MPE $(\downarrow)$ & GED $(\downarrow)$ \\
%\midrule
%Shabani \etal   &  17.64 & 1.02 &\xmark  & \xmark  & 32.24 & 1.10 & \xmark & \xmark  \\  
% TransRaster  &  13.92 & 1.16 & 15.72 & 2.11 & 36.06 & 2.11 &41.93&4.12 \\
% TransVector  &  12.87& 1.08 & 13.97 & 1.99 & 37.72 &   1.93 &42.85 &4.03 \\ 
%Ours  & \bf{4.74\scalebox{0.8}{$\pm 0.6 $}} &\bf{ 0.40\scalebox{0.8}{$\pm 0.0$} }& \bf{5.34\scalebox{0.8}{$\pm 0.8$}}& \bf{0.62\scalebox{0.8}{$\pm 0.0$ }}&\bf{ 17.71\scalebox{0.7}{$\pm0.8$}}&\bf{1.05\scalebox{0.8}{$\pm 0.4$}} &\bf{28.25\scalebox{0.8}{$\pm 0.7$}} & \bf{2.67\scalebox{0.8}{$\pm 0.5 $}}\\ 
\midrule
Shabani \etal  &  29.44 & 1.28 &\xmark  & \xmark  & 36.63 & \bf{1.89} & \xmark & \xmark  \\ 
TransVector  &  36.09 & 1.51 & 46.18 & 2.27 & 40.80 &   2.38 & 53.11 & 6.41 \\ 
Ours  & \bf{8.65} & \bf{0.90}&\bf{10.55}& \bf{0.97 }  & \bf{32.76}& 1.95 & \bf{40.81} & \bf{3.09} \\ 
\bottomrule
\end{tabular}}
\label{table:main}
\end{table*}

%% file: tables/puzzle_main.tex
\begin{table*}[!tb]
\setlength{\tabcolsep}{2pt}
\centering
\caption{CJP and VJP quantitative results with three metrics.
%The evaluation metrics of our proposed method for Jigsaw puzzle solving on Cross-cut (CJP) and Voronoi (VJP) datasets. 
Our method consistently shows superior performance with significant margins under the presence of noise. The second row shows ours without the matching loss. Harel \etal is unable to handle VJP due to the assumptions (See \ref{sec:competing_methods}).
%with the proposed loss function outperforms other baselines in almost all of the experiments. The second row of the methods is our approach without the matching loss.
}
\vspace{2mm}
\scalebox{0.85}{\begin{tabular}{l|ccc|ccc|ccc!{\vrule width 1.5pt}ccc|ccc|ccc!}
  \toprule
     \multicolumn{1}{l}{Dataset} & \multicolumn{9}{c}{Cross-cut Jigsaw Puzzle} & \multicolumn{9}{c}{Voronoi Jigsaw Puzzle} \\
       \cmidrule(lr){1-1}
        \cmidrule(lr){2-10} \cmidrule(lr){11-19}
	\multicolumn{1}{l}{Metric} & \multicolumn{3}{c}{Overlap $(\uparrow)$} & \multicolumn{3}{c}{Precision $(\uparrow)$} & \multicolumn{3}{c}{Recall $(\uparrow)$} & \multicolumn{3}{c}{Overlap $(\uparrow)$} & \multicolumn{3}{c}{Precision $(\uparrow)$} & \multicolumn{3}{c}{Recall $(\uparrow)$} \\ 
 \midrule
 	\multicolumn{1}{l}{Noise Level} & 0 & 1 & 2 & 0 & 1 & 2 & 0 & 1 & 2 & 0 & 1 & 2 & 0 & 1 & 2 & 0 & 1 & 2  \\
\midrule
Harel \etal  &  0.91 & 0.70 & 0.30 & 0.95 & 0.77 & 0.33 & \bf{0.99} & 0.78 & 0.30 & \xmark  & \xmark & \xmark & \xmark  & \xmark & \xmark & \xmark  & \xmark & \xmark \\ 
Ours (w/o $L_{\text{match}}$)  & 0.91 & 0.90 & 0.89 & 0.94 & 0.92 & 0.90 & 0.77 & 0.76 & 0.73 & 0.65 & 0.64 & 0.63 & 0.75 & 0.71 & 0.71 & 0.53 & 0.52 & 0.52 \\ 
Ours  & \bf{0.94} & \textbf{0.94} & \textbf{0.93} & \bf{0.97} & \textbf{0.95} & \textbf{0.93} & 0.91 & \textbf{0.92} & \textbf{0.91} & \bf{0.70} & \bf{0.68} & \bf{0.67} & \bf{0.78} & \bf{0.75} & \bf{0.74} & \bf{0.60} & \bf{0.57} & \bf{0.55} \\ 
\bottomrule
\end{tabular}}
\label{table:mainp}
\end{table*}

%% file: tables/ablation_attention.tex
\begin{table}[!t]
  \setlength{\tabcolsep}{3pt}
  \begin{minipage}{.47\linewidth}
        \centering
        \caption{Contributions of our two attention mechanisms (P-SA, G-SA) and the door matching loss ($L_{\mbox{match}}$).
        %Effect of different components of our method on final performance. 
        Full MagicPlan is used. \checkmark indicates the feature being used. In case of $L_{\mbox{match}}$ ``Doors'' means matching loss has been applied only on door corners  and ``All corners'' means  matching  loss has been applied to all corners including door corners.}
        \scalebox{0.9}{
        \begin{tabular}{c|c|c|c|c}
        \toprule
        P-SA & G-SA & $L_{\scalebox{0.8}{match}}$& MPE $(\downarrow)$ &  GED $(\downarrow)$ \\
        \midrule
        None  & \checkmark&   None  & 48.2 & 4.9 \\ 
        \checkmark& \checkmark&  None    & 43.4 &4.5 \\   
           \checkmark& None    &  All corners &  55.3  &9.4 \\

         None   &\checkmark &  All Corners& 45.1 &  4.6\\   
           \checkmark & \checkmark& All corners &    41.8&   3.6 \\ 
           
        \checkmark&  None    &  Doors &  56.9  &9.3 \\   
      
         None   &\checkmark &  Doors& 45.2 &  4.3\\

         \checkmark & \checkmark& Doors&    40.8&   3.1\\  
        \bottomrule
        \end{tabular}}
        \label{table:main_ab}
  \end{minipage}%
  \hspace{0.3cm}
  \begin{minipage}{.47\linewidth}
        \centering
        \caption{Effects of the room-type (R-type) and the Door information. Full MagicPlan is used. \checkmark indicates the information being used. When a room-type is not used, we set a zero vector as a room-type one-hot vector, when room type is noisy we assign each room with a random room type . When the door information is not used, we do not pass the door-corner nodes to the network.}
        \scalebox{0.9}{
        \begin{tabular}{c|c|c|c!{\vrule width 1.5pt}c|c}
        \toprule
        \multicolumn{2}{c}{Train}  & \multicolumn{2}{c!{\vrule width 1.5pt}}{Test} & \multirow{2}{*}{MPE$(\downarrow)$} & \multirow{2}{*}{GED $(\downarrow)$}\\
          \cmidrule(lr){1-2}  \cmidrule(lr){3-4}  R-Type & Door  &   R-Type   & Door & &\\ \hline
              \checkmark & \checkmark &  None   & \checkmark &  48.4& 3.8\\\
               
                    None & \checkmark & None    & \checkmark &  47.4& 3.6\\
                     \checkmark & \checkmark & Noisy    & \checkmark &  49.7& 4.0\\

               Noisy & \checkmark & Noisy    & \checkmark &  49.3& 3.9\\
               
           \checkmark & \checkmark & \checkmark &   None  &46.9 & 5.2 \\
          \checkmark &  None          & \checkmark &  None   &  45.5 &5.2 \\
          
          \checkmark &  \checkmark &     \checkmark & \checkmark  & 40.8 &3.1\\
        \bottomrule
        \end{tabular}}
        \label{table:door}
  \end{minipage} 
\end{table}

%% file: sections/8_conclusion.tex
\subsection{Conclusion}
This paper introduced an end-to-end neural architecture for spatial puzzle solving tasks. 
The proposed approach is faster, robust to data corruptions, end-to-end, and far superior to existing methods in all the metrics in the variety of tasks, namely Cross-cut jigsaw puzzle (pictorial and apictorial), Voronoi jigsaw puzzle, and room layout arrangement.
Despite numerous advantages, our approach faces certain limitations. The primary challenge is its substantial need for training data, which limits our ability to process smaller datasets. 
For the room layout arrangement task, the utilization of raw image information could further enhance performance as discussed in~\cite{amin,zillow}. However, this would significantly increase the amount of data transfer from a mobile device, where on-device processing would become desirable. 
One key future work is the development of data-efficient (at training) and computation-efficient (at testing) neural architectures~\cite{Kitaev2020Reformer:}.
%
%, could significantly enhance the performance of our model. However, this approach would necessitate the minimization of data transfer, possibly through the application of on-device processing.
%
To our knowledge, this paper is the first to demonstrate that Diffusion Models, generally regarded as powerful generative models, are also effective in solving various challenging spatial arrangement tasks. The paper has the potential to
motivate other researchers to further expand the applicability of emerging Diffusion Models, moving beyond content generation and into a myriad of other tasks.

%In conclusion, while our approach represents a significant advancement in spatial puzzle-solving tasks, there remains ample opportunity for future enhancements and broader applications.

%However, the end-to-end design with a powerful neural architecture requires more training data, prohibiting us from processing some other smaller datasets. One future work is the integration with more data efficient neural architecture~\cite{Kitaev2020Reformer:}. Another future work is the utilization of image information~\cite{amin,Zillow}, while minimizing the amount of data transfer by utilizing on-device processing. This method has a potential of opening new doors on the use of emerging diffusion models in a much broader set of tasks than content generation.

%% file: sections/sups-method.tex
 \section{Methods details of our system and competing methods}
 \label{sec:arc}

We benefit from Transformers in our task in two ways. First, Transformers provide the capability of processing sequences with different lengths, which we use to process different number of room layouts/corners in the houses. Second, we utilize the self-attention module of Transformers to create optimal interaction and information-sharing among input tokens. These two features make Transformers an ideal backbone for our model.
Our method uses six Transformer encoder blocks, and attention in each block has four heads. We also use an MLP For converting 256D Transformer output to rotation and position (4D). To keep the experiments fair, we use the same architecture for our transformer baselines as much as possible. In the following, we provide details corresponding to each of the baselines.

\mypara{Transformer with a raster representation} (TransRaster) uses the raster images to represent the input room layouts/types and the output room positions. Note that this baseline does not handle rotations as explained below.
%nd room types representation as the input. 
An input room layout is represented as a 20-channel $256 \times 256$ semantic segmentation image, where there are 20 room/door types. The room center is aligned with the center of an image.
%mask per room where each channel represent one room type and doors on channel 25; 
An output room position is represented as a $256 \times 256$ room occupancy image, which is ideally a translated version of the input room segmentation image at the correct room location.
Given an output room occupancy image, we perform an exhaustive search over the possible room translations and find one with the most overlap between the occupancy image and the translated room segmentation image.~\footnote{We could expand the search space with possible room rotations, but rooms are often symmetric. To be simple, we use this baseline only for experiments when ground-truth rotations are given.}
We use VisionTransformer~\cite{vit} with a CNN decoder that takes a set of input room segmentation images and produces a set of room occupancy images.

In the other word TransRaster uses an encoder part of U-Net, which has 8 down-sampling blocks, converting each input room layout to a feature map of dimension 512. Each feature map (corresponding to a room layout) will become one input token for the Transformer. Input sequence’s length is equal to the number of rooms in a house, and information is shared among different rooms. We use six Transformer encoder blocks, and attention in each block has four heads.
We pass the output of Transformer to a U-Net up-sampling model with eight Up-sampling blocks to change the dimension from 512 to $256 \times 256$. 

% After Transformer encoder blocks, we use  U-Net up-sampling model with eight Up-sampling blocks to change the dimension from 512 to $256 \times 256$. 

\mypara{Transformer with a vector representation} uses the same backbone as our method; a linear layer converts the 28D input vector (i.e., 2 for the original corner coordinate and 20 for the room/door type one-hot vector) to the 256D feature map, six Transformer encoder blocks, and the attention in each block have four heads. We also use an MLP to convert 256D output embedding to 4D output.

 \mypara{Diffusion model, one room per node} encodes each node as corresponding to a room instead of a corner in the room. To ease the implementation, we set the maximum number of nodes per room to 20 and we pad extra nodes when the room has less than 20 nodes with 0. We flatten the conditions per room and then use a linear layer to convert it to a 256D embedding vector. Each feature map represents a room and an input token for Transformer, we use the same Transformer as our method. After the Transformer blocks, a linear layer converts 256D output to 4D (i.e., 2 for the position and 2 for the rotation).
 
 \mypara{Shabani \etal}~\cite{amin} takes the input layout of each room with the resolution of $256\times 256$ with the same number of channels as the number of room types to pass each pixel as a one-hot vector of the corresponding room type. We use the same model as \cite{amin} and change the number of input channels to 11  for RPLAN and 20 for MagicPlan. To generate the arrangement candidates, we use the given room layouts of our dataset to connect doors, while we also use overlap filtering to reduce the number of candidates.
 Note that our datasets is significantly larger than the one in \cite{amin}, enabling us to randomly select a positive or a negative candidate in each iteration and therefore remove the class imbalance weight used in \cite{amin}. During the training for each house, we randomly select a GT with the label 1 or a faulty candidate with the label [0, 1) based on the number of mismatched doors. During the test, we pass all the possible candidates of each house and select the candidate with the highest score as the final prediction.

\mypara{Harel \etal}~\cite{harel2021crossing} proposed a two-step algorithm for CJP. Their method considers two types of constraints to find plausible mates based on the length and angle of different pairs of connections. By estimating the matings hierarchically using these constraints, they approach the problem of finding positions as a multi-body spring-mass system. We utilize the authors' provided implementation\footnote{\href{https://icvl.cs.bgu.ac.il/polygonal-puzzle-solving/}{https://icvl.cs.bgu.ac.il/polygonal-puzzle-solving/}} for comparison with our method. With a test dataset of 1000 crossing-cut puzzles, we restrict the running time of the spring-system algorithm to 2 minutes per puzzle. Furthermore, unlike the provided implementation, we also consider failure cases in the metrics. Regarding the pictorial case, the authors score a candidate mating by extrapolating the images of puzzle pieces and considering the difference of the mean color value on the edges. We do not impose any time limit as we evaluate only on 20 samples.

%% file: sections/sups-dataset.tex
\section{Datasets Details and Preprocessing}
 \label{sec:dataset}

\begin{table*}[!htb]
    \centering
    \begin{tabular}{l!{\vrule width 1.5pt}c|c|c|c|c|c|c|c|c|}
    \toprule
       Room Type  &3 & 4 & 5& 6& 7& 8& 9& 10& All  \\
         \midrule
    Master bedroom  &0.20 & 0.26  & 0.29 &0.32 &0.39  & 0.49 & 0.49 & 0.53 & 0.34  \\
   Living room   &  0.52  & 0.56  & 0.59 & 0.65 & 0.71 &0.75  &0.79  & 0.79 &  0.65 \\
   Kitchen   & 0.42 &  0.47 & 0.52 &  0.59 & 0.68 & 0.71 & 0.76 & 0.79 & 0.59  \\
   Bathroom &  0.54 &  0.70 & 0.85 &0.96    & 1.12 & 1.28 & 1.33 & 1.47 &  0.96 \\
    
    Toilet & 0.07  &  0.11 & 0.15 &0.22 & 0.22 &0.25  & 0.27 &0.26  &  0.18 \\
    
 Corridor & 0.07  &  0.12 &  0.15&0.19 & 0.25 & 0.32 &0.37  & 0.45 &  0.21 \\
    
    Closet &  0.13 &  0.18 & 0.22& 0.32  &0.48  & 0.68 & 0.92 &1.22  &  0.41 \\
    
    Hall & 0.35  &  0.55 & 0.68 &0.77 & 0.85 & 0.91 & 0.98 & 1.08 &  0.73 \\
     Laundry room  & 0.05  &0.05  & 0.05&0.07  & 0.10 & 0.13 & 0.18 & 0.23 &0.09   \\
     Bedroom   & 0.34 & 0.69  &1.08  &1.32 &1.51  & 1.74 & 1.93 &2.10  &  1.23 \\
   Balcony &  0.05 &  0.08 &0.16  & 0.32  &0.40  &0.48  & 0.56 & 0.64 &  0.29 \\
     Dining room & 0.13  &  0.12 &  0.12&0.14 &0.17  & 0.19 & 0.22 & 0.25 &  0.15 \\
    Private office & 0.00  & 0.01  &0.02  &0.05 & 0.05 &0.07  & 0.08 &  0.10&  0.05 \\
     Den &  0.05  & 0.05  & 0.06 &0.7 & 0.09 & 0.11 & 0.12 & 0.12 & 0.08  \\
    Storage  & 0.00  &  0.01 &  0.01& 0.02 &0.02  & 0.02 & 0.03 &0.03  &  0.02 \\
    Others &  0.00 &  0.01 &  0.01& 0.02& 0.03 & 0.04 &0.04  &0.07  &  0.03 \\
   Doors  & 2.84&3.82 &4.82 &5.43&6.92 & 8.02&6.10&10.18 & 5.90 \\
   \bottomrule
    \end{tabular}
    \caption{MagicPlan dataset consists of floorplans with 3 to 10 rooms. The table shows average number of rooms with a specific room type based on the total number of rooms in the house.}
    \label{tab:dataset}
\end{table*}

We normalize the puzzles/floorplans for each task and dataset by scaling them to fit within a $1\times1$ square, and we also resize all corresponding images to dimensions of $256\times 256$ in the case of pictorial CJP. While this normalization process does not introduce any essential additional information during testing in CJP and VJP, it could potentially enables the network to cheat in RLA, as the longer extent of arranged floorplans is always fixed to $1$. To address this issue, during testing in RLA, we apply a random scaling factor in the range of [0.8, 1.0] to the room shapes of each house. In the subsequent sections, we provide a detailed description of each dataset. In the following, we provide additional statistics for our floorplan datasets.

The Voronoi Jigsaw Puzzle dataset consists of 200k training puzzles and 1k testing puzzles. These puzzles were created by randomly selecting 3 to 15 points, The individual pieces of the puzzles were obtained by extracting the Voronoi cells corresponding to these points. There are 1,066, 14,033, 23,715, 20,279, 16,073, 15,235, 16,428, 15,018, 16,318, 15,096, 16,487, 16,233, and 15,670 puzzles with 3, 4, 5, 6, 7, 8, 9, 10, 11, 12, 13, 14, and 15 pieces respectively. Each piece has a minimum, maximum, and average of 3, 20, and 4.51 corners respectively. The minimum, maximum, and average number of corners per puzzle are 10, 93, and 42.24.

Cross-cut Jigsaw Puzzle (CJP) are consist of  100k training  and 1k testing puzzles, where each one were generated using ~\cite{harel2021crossing} method which generate a convex polygon and cuts it by 3 to 5 lines. There are 1719, 6046, 15854, 14521, 6905, 10929, 12065, 8361, 6521, 8192, 7663, 4642, 1508, 73, and 1 puzzles with 3, to 18 pieces respectively. Each piece has a minimum, maximum, and average of 3, 13, and 4.47 corners respectively. The minimum, maximum, and average number of corners per puzzle are 16, 76, and 41.99.

MagicPlan dataset consists of roughly 98K houses/apartments, which we divide into 93K training and 5K testing samples. The number of rooms in a house ranges from 3 to 10. Concretely, 11661, 16322, 19171, 17582, 13200, 9649, 6780, and 4415 houses contain 3, 4, 5, 6, 7, 8, 9, and 10 rooms, respectively. The minimum and maximum numbers of corners in a house are 12 and 182. 
Table~\ref{tab:dataset} shows average number of rooms with a specific room type based on the total number of rooms in the house.

In the RPLAN dataset, we divide 60K samples in RPLAN to 55K train and 5K test. The number of rooms in a house ranges from 3 to 8. Concretely 99, 582, 5083, 19551, 21921, and 13235 houses contain 3, 4, 5, 6, 7, and 8 rooms, respectively.

%% file: sections/sups-exps.tex
\section{Additional ablation studies }
\label{sec:abs}

\subsection{Additional ablation studies on room layout arrangement}

\begin{table}[!tbh]
\setlength{\tabcolsep}{3pt}
\centering
\caption{Main quantitative results with two metrics: 
%comparison of our method with the baselines on different settings using
Positional Error (MPE) and Graph Editing Distance (GED).
% as the metrics. 
This table show a case where the ground-truth rotations are given, as TransRaster baseline cannot handle rotations.
Small RPLAN (resp. Small MagicPlan) is a subset of the
corresponding full dataset, consisting of houses with at most 6 rooms. The small datasets are created for Shabani \etal, which is not scalable to many rooms.
Our method is stochastic and shows both the mean and the standard deviation.
%which 
%We compare in four scenarios based on the number of rooms and rotation. Small RPLAN and Small JigsawPlan have only houses with up to 6 rooms.  We compare the methods with fixed rotation in the \textbf{top} and without fixing the rotation in the \textbf{bottom} where Raster Transformer fails.
%The proposed method outperforms the baselines in all of the experiments except Graph Distance for Small JigsawPlan which it gets comparable results with the state-of-the-art.
}
\scalebox{0.95}{\begin{tabular}{l|c|c|c|c!{\vrule width 1.5pt}c|c|c|c!}
		\toprule
     \multicolumn{1}{l}{Dataset} & \multicolumn{2}{c}{Small RPLAN} & \multicolumn{2}{c}{Full RPLAN} & \multicolumn{2}{c}{Small MagicPlan }& \multicolumn{2}{c}{Full JigsawPlan}\\
       \cmidrule(lr){1-1}
        \cmidrule(lr){2-3} \cmidrule(lr){4-5} \cmidrule(lr){6-7} \cmidrule(lr){8-9} 
	\multicolumn{1}{l}{Metric} & MPE $(\downarrow)$ & GED $(\downarrow)$ & MPE $(\downarrow)$ & GED $(\downarrow)$ & MPE $(\downarrow)$ & GED $(\downarrow)$ & MPE $(\downarrow)$ & GED $(\downarrow)$ \\
\midrule

Shabani \etal   &  17.6 & 1.0 &\xmark  & \xmark  & 32.2 & 1.1 & \xmark & \xmark  \\  
 TransRaster  &  13.9 & 1.2 & 15.7 & 2.1 & 36.1 & 2.1 &41.9&4.1 \\
 TransVector  &  12.9& 1.1 & 13.9 & 2.0 & 37.7 &   1.9 &42.8 &4.0 \\ 
Ours  & \bf{4.6\scalebox{0.8}{$\pm 0.7 $}} &\bf{ 0.4\scalebox{0.8}{$\pm 0.0$} }& \bf{5.4\scalebox{0.8}{$\pm 0.7$}}& \bf{0.6\scalebox{0.8}{$\pm 0.0 $ }}&\bf{ 17.5\scalebox{0.7}{$\pm0.8$}}&\bf{1.0\scalebox{0.8}{$\pm 0.4$}} &\bf{27.9\scalebox{0.8}{$\pm 0.7$}} & \bf{2.7\scalebox{0.8}{$\pm 0.5 $}} \\ \bottomrule

\end{tabular}}
\end{table}

\begin{figure}[!tbh]
  \begin{minipage}[t]{.44\textwidth}
    \centering
    \includegraphics[width=\linewidth]{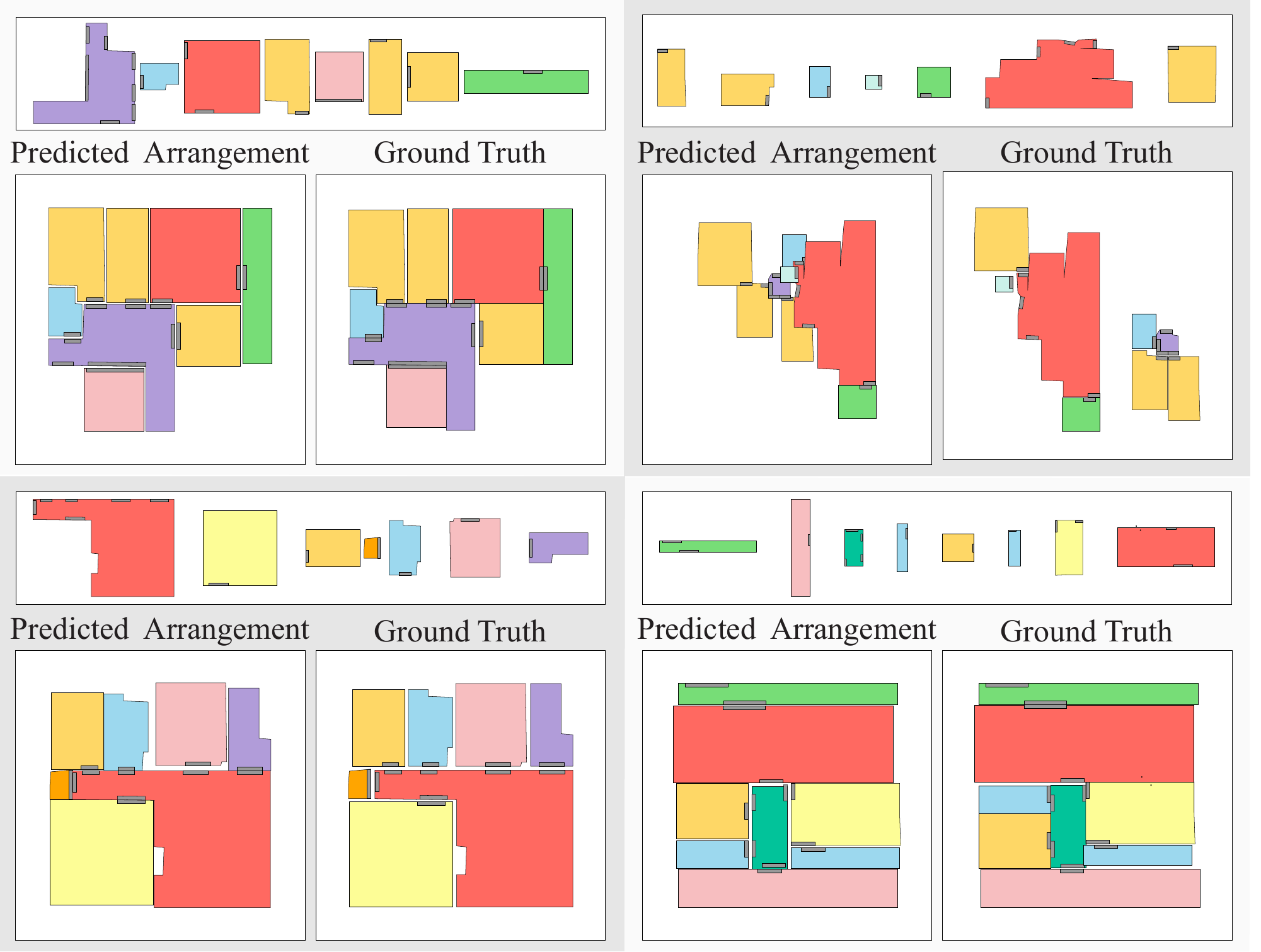}
    \caption{RLA arrangement results with
Full MagicPlan dataset. On left two successful cases, on right two failed cases,.  Our failures are often attributed to 1) Rare building architecture (top-right) and 2) Inherent ambiguity (bottom-right ), whose tasks are challenging even for humans.}
    \label{fig:s-f-case}
   \end{minipage}
\hspace{0cm} % Add horizontal space here
\begin{minipage}[t]{.54\textwidth}
    \centering
    \includegraphics[width=\linewidth]{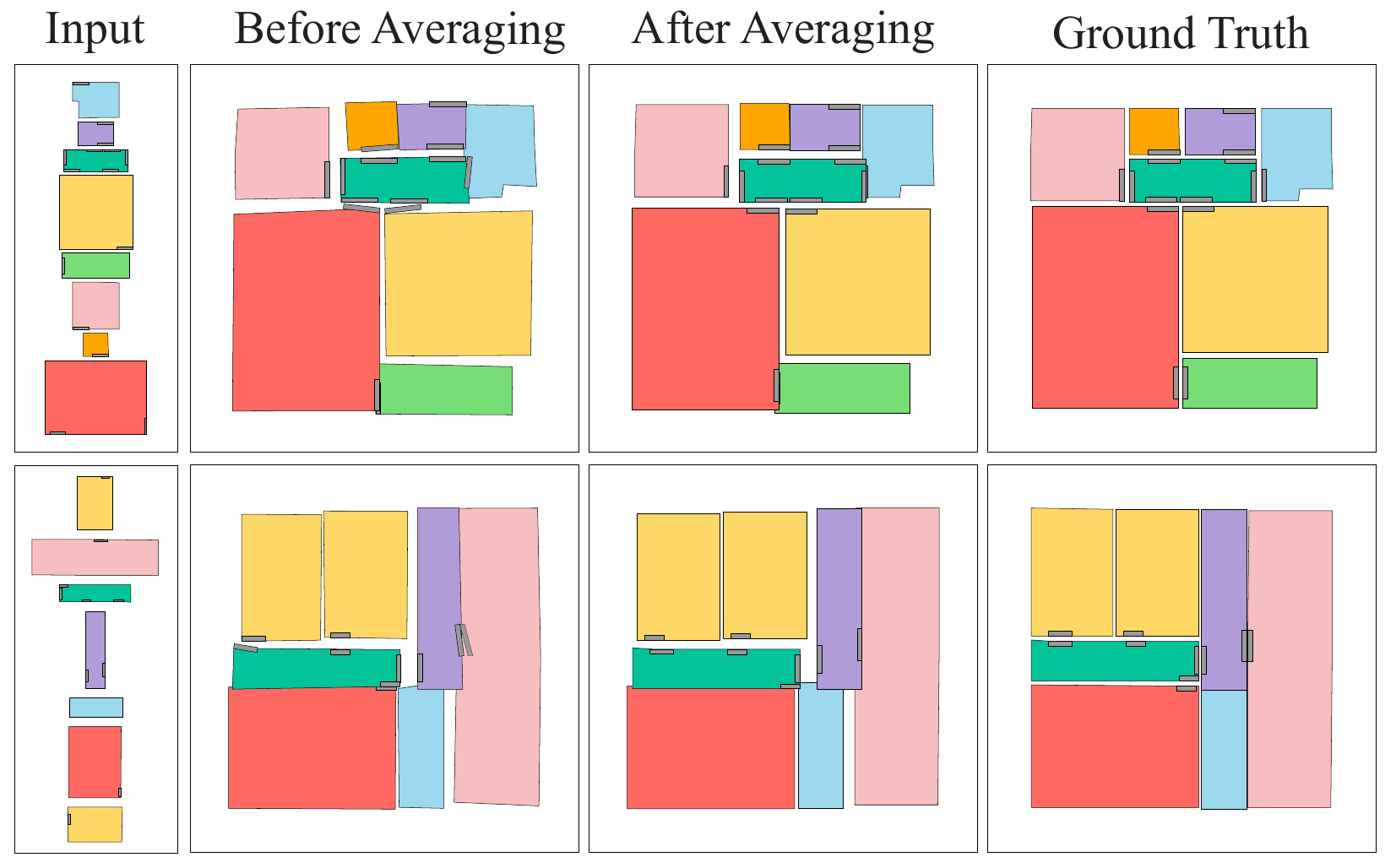} 
    \caption{
    % The final room arrangement before and after averaging. Our diffusion model estimates a room position/rotation at each room corner, which may not be consistent in a room. The final arrangement is obtained by taking the average position (the majority vote for rotation) within each room. The second (resp. third) column shows the results before (resp. after) the averaging. 
    The final room arrangement before and after averaging. Our diffusion model estimates a room position/rotation at each room corner, which may not be consistent in a room. The final arrangement is obtained by taking the average position and rotation within each room.
    }
  \label{fig:ensemble}
  \end{minipage}
\end{figure}

\begin{figure}[!h]
    \centering
 \noindent\includegraphics[width=0.99\textwidth]{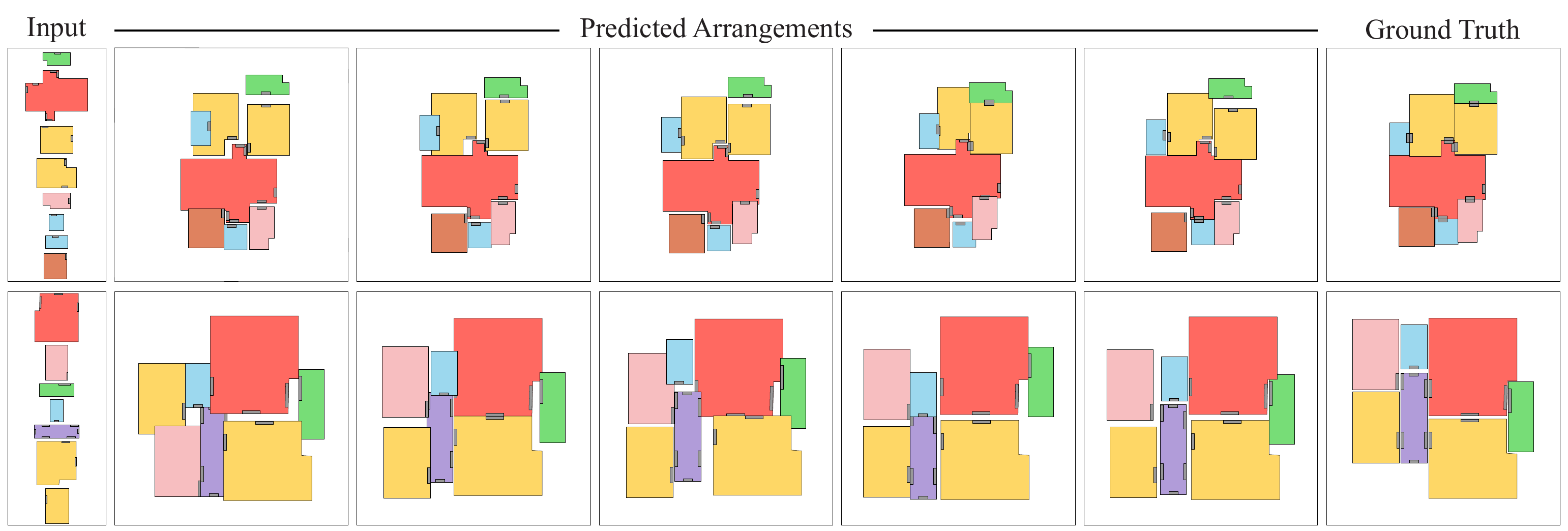} 
    \caption{A diffusion model is stochastic and produces a different result every time. The middle rows show five different pose estimation results. The top (resp. bottom) is from Full RPLAN (resp. Full MagicPlan) dataset.
    }
    \label{fig:diffrent_run}
\end{figure}

Figure~\ref{fig:ensemble} shows the raw estimated position information at each room/door corner before the room-wise averaging. Since the ground-truth has the same pose parameters for all corners in a room/door, the network learns to produce consistent parameters. Figure~\ref{fig:diffrent_run} shows five pose estimation results by our system while varying the initial noise $x_T$.
%To ensure that our method predicts reasonably consistent regardless of the initial random sample $x_T$, we visualize five random predictions of our method using a fixed input condition in Fig.~\ref{fig:diffrent_run}. 
While there are minor differences, the overall room arrangements are similar and close to the ground-truth, indicating that the Diffusion model is capable of producing consistent results given enough constraints as a pose estimation system, as opposed to a generative model whose original goal is to create a diverse set of answers.

\begin{figure}[!tbh]
    \centering
 \noindent\includegraphics[width=
 0.99\textwidth]{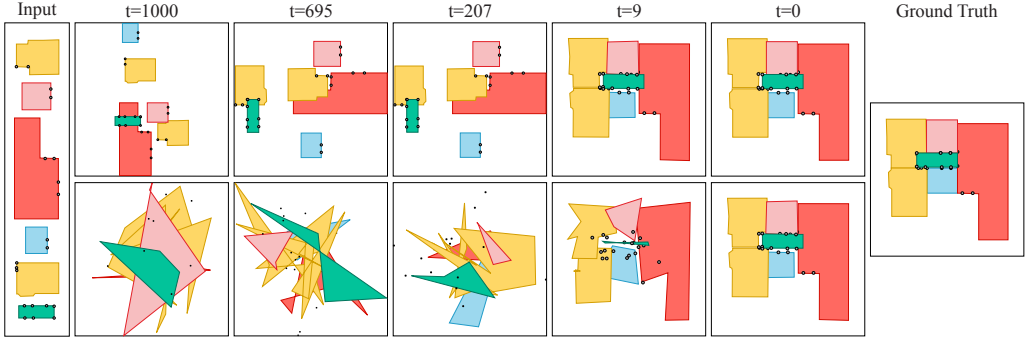} 
    \caption{Visualization of predicted layouts at step ``t''s. At t=1000, position parameters at each corner are initialized by a Gaussian noise, and at t=0, there is the final predicted layout. The top row shows the predicted layout without averaging/voting, and the bottom row shows with averaging/voting. To make it more clear, we show doors by their corners.}
    \label{fig:steps}
\end{figure}

  \begin{figure}[!th]
    \centering
 \noindent\includegraphics[width=\textwidth]{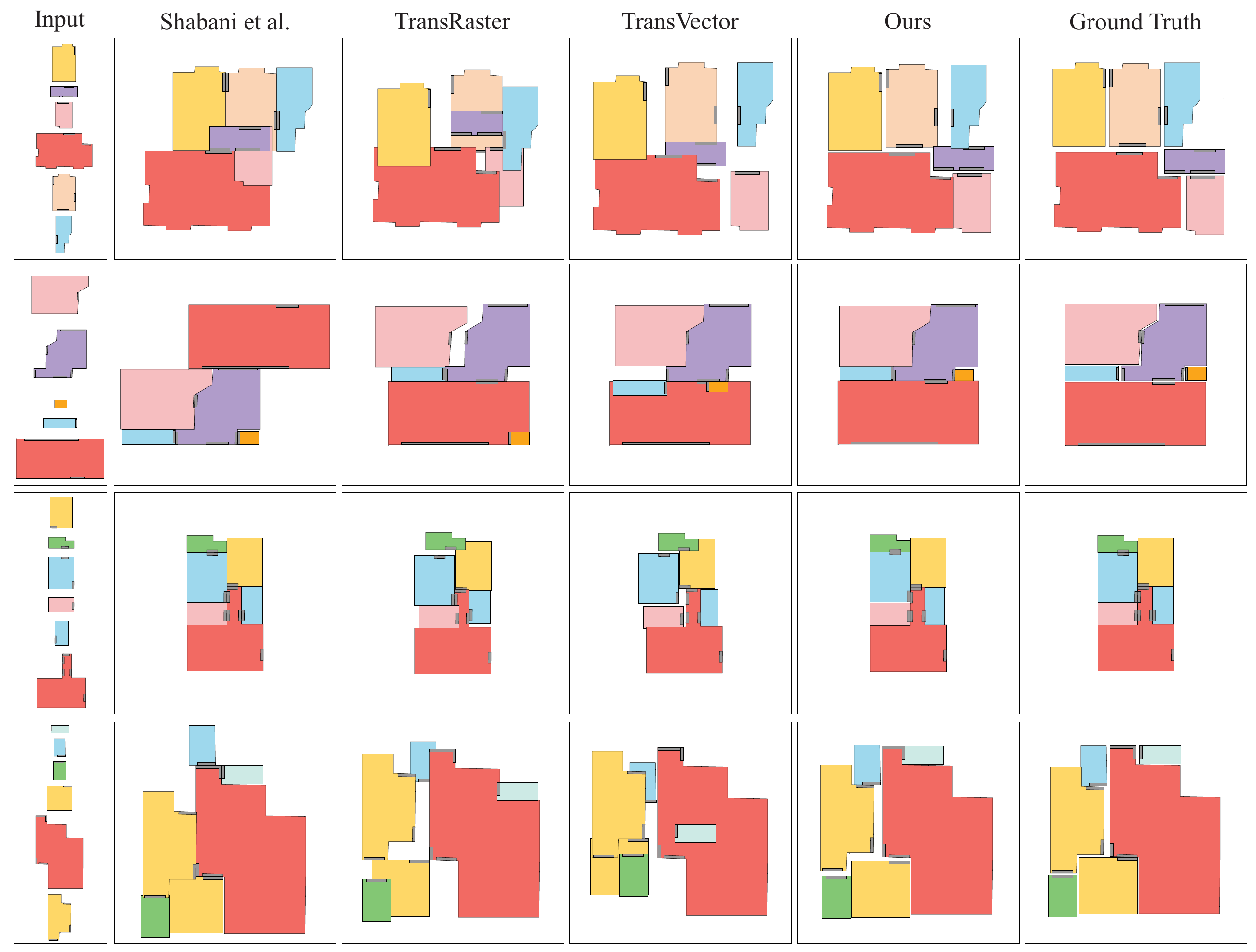} 
    \caption{Qualitative evaluations of our approach against the three competing methods. The top two rows are from Small MagicPlan. The bottom two rows is from Small RPLAN. The GT rotations are given for all the cases to enable comparisons with all the methods.}
   
%    comparison of our method with the all of the baselines on small versions of MagicPlan (first two rows) and RPLAN (last row) with fixed rotations. Our methods predicts the correct room positions in almost all of the samples while the other baselines mostly fail.

    \label{fig:main}
\end{figure}
  \begin{figure}[!tbh]
    \centering
 \noindent\includegraphics[width=0.99\textwidth]{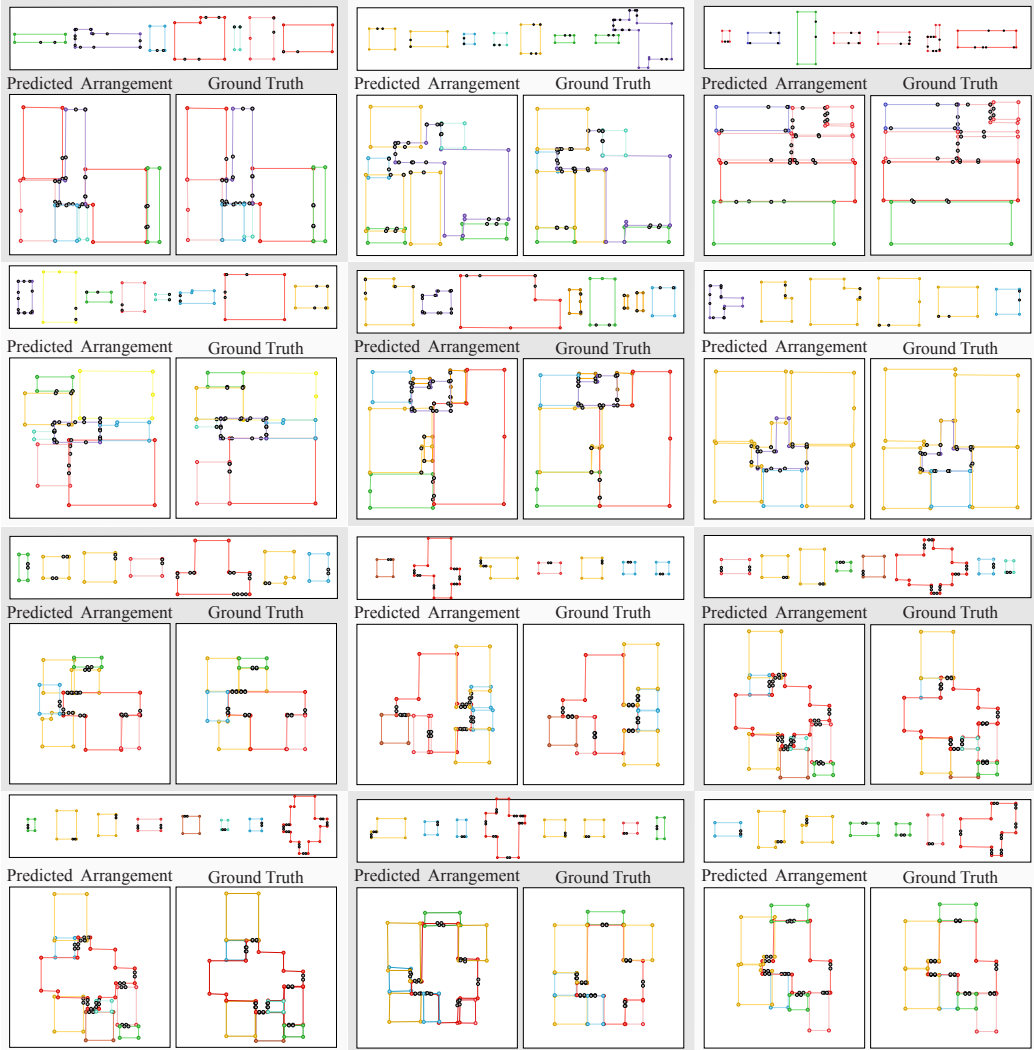} 
    \caption{Qualitative evaluations of our method for Full MagicPlan dataset without GT rotations top two rows, and Full RPLAN dataset without GT rotations bottom two rows. We show edges and corners here to show overlaps and noisy annotations more clear.}
    %
%Additional examples of predictions by our method, including the correct predictions (Top) and failure cases (Bottom). Considering most of the floorplans in our dataset are connected, our method tends to generate connected floorplans. The other errors are caused by strange floorplan shape or multiple possible choices which in both the proposed method generates reasonable results.}
    \label{fig:s-f}
\end{figure}

\subsubsection{Additional ablation on RPLAN}
The main paper shows the ablation studies on the MagicPlan dataset in case of room layout arrangment task. This part of supplementary will present the same study results on RPLAN dataset.
%In the main paper Full MagicPlan dataset were used for ablation studies,here We repeat some of those studies on RPLAN dataset. 
Table~\ref{table:main_ab} shows the impact of our attention module and door matching loss on performance and  Table~\ref{table:door} shows the impact of noise in the room type and door detection on our performance, although there is a performance drop, our method still works better than the competing methods.

 \begin{table}[!tbh]

\begin{minipage}{.47\linewidth}
        \centering
\caption{Co attention mechanisms (P-SA, G-SA) and the door matching loss ($L_{\mbox{match}}$).
        %Effect of different components of our method on final performance. 
        Full RPLAN is used. \checkmark indicates the feature being used. In case of $L_{\mbox{match}}$ ``Doors'' means matching loss has been applied only on door corners  and ``All corners'' means  matching  loss has been applied to all corners including door corners.}
\scalebox{0.8}{\begin{tabular}{c|c|c!{\vrule width 1.5pt}c|c}
\toprule
P-SA & G-SA & $L_{\scalebox{0.8}{match}}$& MPE $(\downarrow)$ &  GED $(\downarrow)$ \\
\midrule
 & \checkmark&    & 25.6& 1.6\\ 
\checkmark& \checkmark& & 24.2&1.5 \\   
\checkmark&    & Doors & 36.9 &2.4 \\   
   &\checkmark &  Doors& 22.1 &  1.1 \\   
   \checkmark & \checkmark& All Corners&    10.7 &   0.9 \\  
 \checkmark & \checkmark& Doors&    10.5 &   0.9 \\  
\bottomrule
\end{tabular}}
\label{table:main_ab}
\end{minipage}\hspace{0.3cm}
  \begin{minipage}{.47\linewidth}
\centering
\caption{Effects of the room-type (R-type) and the Door information. Full RPLAN is used. \checkmark indicates the information being used. When a room-type is not used, we set a zero vector as a room-type one-hot vector. When the door information is not used, we do not pass the door-corner nodes to the network.}
\scalebox{0.8}{\begin{tabular}{c|c|c|c!{\vrule width 1.5pt}c|c}
\toprule
\multicolumn{2}{c}{Train}  & \multicolumn{2}{c!{\vrule width 1.5pt}}{Test} & \multirow{2}{*}{MPE$(\downarrow)$} & \multirow{2}{*}{GED $(\downarrow)$}\\
  \cmidrule(lr){1-2}  \cmidrule(lr){3-4}  R-Type & Door  &   R-Type   & Door & &\\ \hline
      \checkmark & \checkmark &     & \checkmark &17.3& 1.4\\
             & \checkmark &     & \checkmark &  16.4& 1.5 \\
   \checkmark & \checkmark & \checkmark &     &15.1 & 1.9 \\
  \checkmark &            & \checkmark &     & 14.3 &1.9 \\
  \checkmark &  \checkmark &     \checkmark & \checkmark  & 10.5 &0.9\\
\bottomrule
\end{tabular}
\label{table:door}}
\end{minipage}
\end{table}

\subsection{Additional ablation studies on puzzle solving}
\begin{figure}[!hbt]
   \centering
    \includegraphics[width=0.95\textwidth]{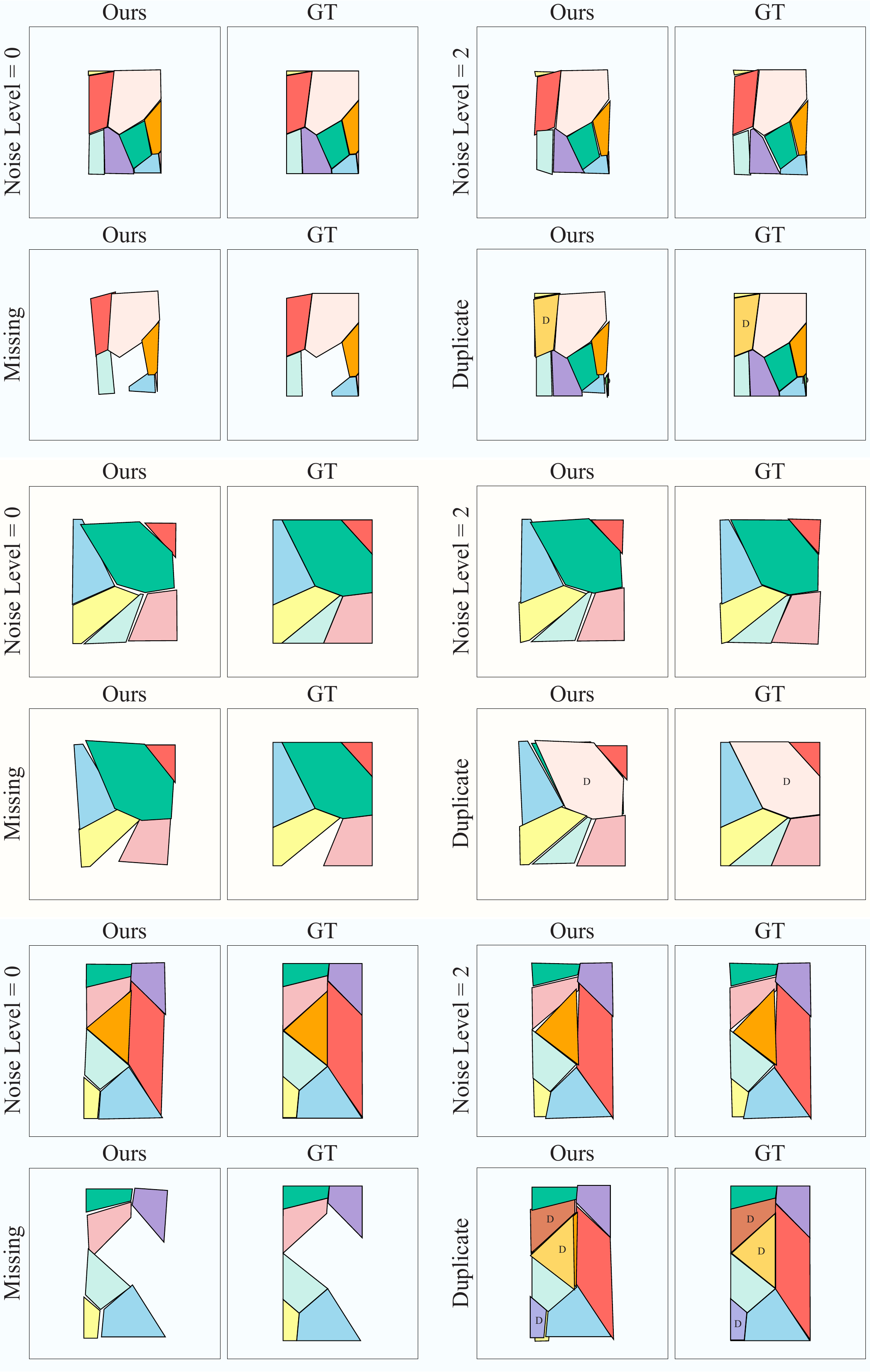}
    \caption{Additional qualitative results of Voronoi jigsaw puzzle are presented in four different setups: 1) No noise, 2) Noise level 2, 3) Missing piece, and 4) Duplicate piece (D indicates the duplicated pieces). }
    \label{fig:qualitative_voronoi}
\end{figure}

\begin{figure}
   \centering
    \includegraphics[width=.92\textwidth]{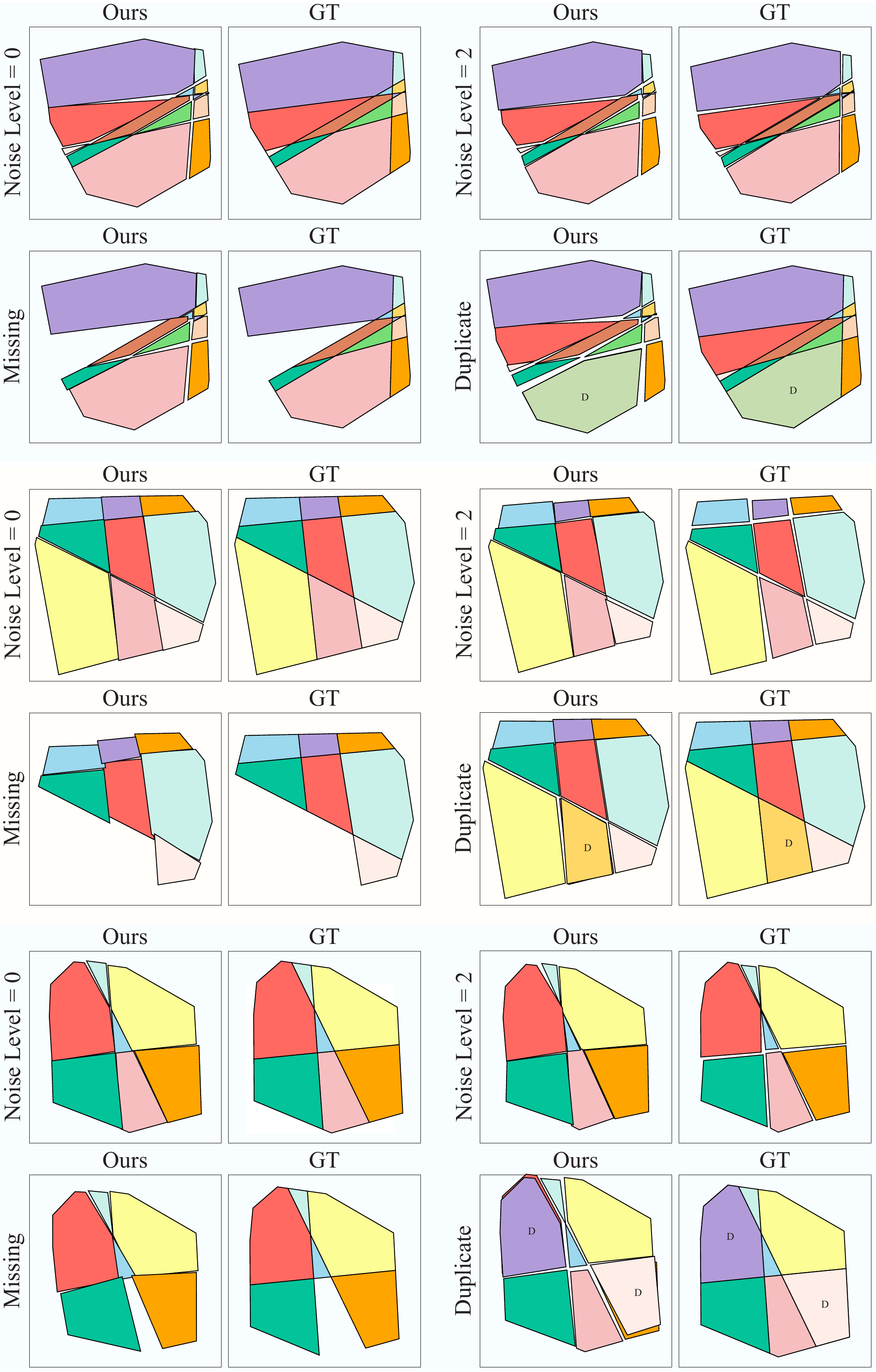}
    \caption{Additional qualitative results of Cross-cut jigsaw puzzle are presented in four different setups: 1) No noise, 2) Noise level 2, 3) Missing piece, and 4) Duplicate piece (D indicates the duplicated pieces).}
    \label{fig:qualitative_crosscut}
\end{figure}

% In this section we show extra qualitative results.
% Furthermore Table~\ref{tab:miss-extra} presents our method performance in case of duplicate or missing pieces.
We have provided additional qualitative results of our method in Figure~\ref{fig:qualitative_voronoi} and Figure~\ref{fig:qualitative_crosscut} including noisy samples or samples with missing or duplicate pieces. In case of missing and duplicate experiment, we repeat (remove) each piece with a probability of $10\%$. Table~\ref{tab:miss-extra} presents the evaluation metrics of missing and duplicate experiments.

\subsubsection{Pictorial Cross-cut Jigsaw Puzzle}
To enhance the integration of image information into our pictorial puzzle diffusion models, we employed a two-step approach. Firstly, we pretrained an auto-encoder utilizing the puzzle pieces. This auto-encoder served as the image embedder for our diffusion model, enabling the conversion of each puzzle into a compact 128D feature vector.

The pretraining process involved training the model to downsample an input image of dimensions $3\times256\times256$ to a compressed representation of size $32\times2\times2$ within the encoder, and subsequently reconstructing the original image size in the decoder. We employed the mean squared error (MSE) loss function during training. However, to focus our model's attention on learning the texture features, given that the diffusion model already captured the geometry features, we applied the loss function exclusively to the pixels within the puzzle piece.

By adopting this selective application of the loss function, we prioritize the acquisition of texture-based details, as the geometric characteristics are already embedded within the diffusion model. 

\begin{figure}
   \centering
    \includegraphics[width=0.9\textwidth]{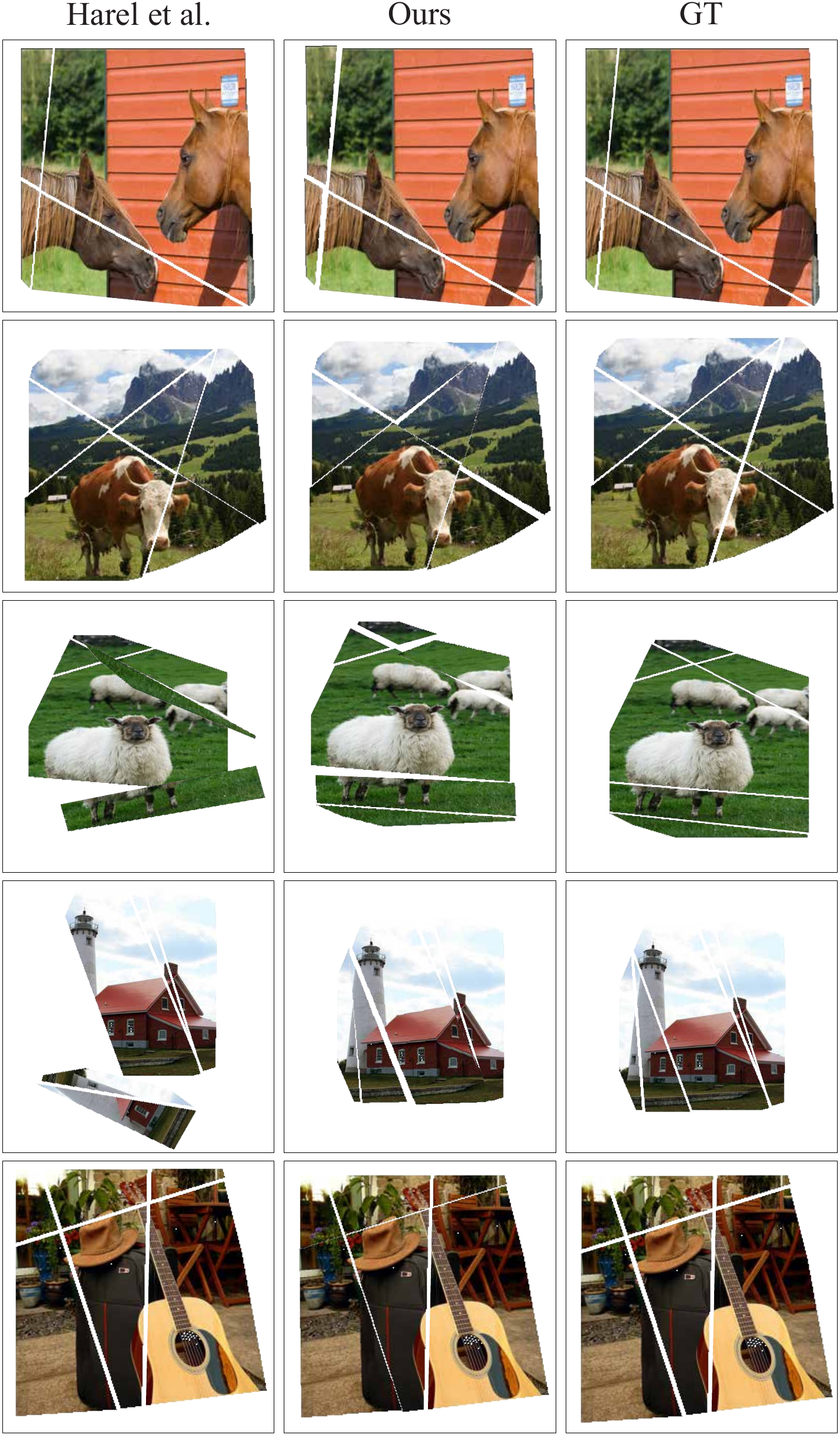}
    \caption{Additional qualitative results of pictorial Cross-cut jigsaw puzzle compared to Harel \etal~\cite{harel2021crossing}.}
    \label{fig:qualitative_crosscut_pic}
\end{figure}

Quantitatively, we also evaluated our method on the full Cross-cut dataset to measure the effectiveness of the pictorial information compared to apictorial scenario. We found that the model converges faster when using pictorial information while it achieves slightly better overlap score of 0.9417 compared to 0.9398 in apictorial scenario. Figure~\ref{fig:qualitative_crosscut_pic} shows additional qualitative results of our method compared to Harel \etal~\cite{harel2021crossing}. 

% \begin{table}[!tbh]
%     \centering
%        \caption{pictorial vs apictorial}
%     \begin{tabular}{c|ccc}
% \toprule
%   Method & {Overlap $(\uparrow)$} & {Precision $(\uparrow)$}&  {Recall $(\uparrow)$} \\
%   \midrule
%   APictorial & 0.91 \\
%   Pictorial & 94.16 & 96.76 & 90.49 \\
%            \bottomrule
%     \end{tabular}
%     \label{tab:pictvsapict}
% \end{table}

\begin{table}[!h]
\setlength{\tabcolsep}{2.6pt}
    \centering
       \caption{Effects of the Missing or Duplicate pieces in puzzle solving problem. \checkmark indicates it if missing or duplicate piece were presented during test time. In training time we do not have duplicate or missing piece presented to show our model robustness to unseen noise during test. }
    \begin{tabular}{ c c |ccc |cc c }
    \toprule
       \multicolumn{2}{c}{}&   \multicolumn{3}{c}{Cross-cut } & \multicolumn{3}{c}{Voronoi} \\    \cmidrule(lr){1-2} \cmidrule(lr){3-5} \cmidrule(lr){6-8}
  Missing& \multicolumn{1}{c|}{Duplicate} & {Overlap $(\uparrow)$} & {Precision 
  $(\uparrow)$} & {Recall} $(\uparrow)$  &   {Overlap $(\uparrow)$} & {Precision $(\uparrow)$}& {Recall}  $(\uparrow)$ \\\hline
      
        \checkmark & - &         0.88 & 0.92 & 0.82 & 0.68 & 0.68 & 0.56 \\
          - & \checkmark  &       0.92 & 0.97 & 0.88 & 0.67 & 0.71 & 0.57 \\
           - & - &                 0.94 & 0.97 & 0.91 & 0.70 & 0.78 & 0.60 \\
           \bottomrule
    \end{tabular}
 
    \label{tab:miss-extra}
\end{table}

%% file: neu.bbl

\begin{thebibliography}{00}

%%% the Web address.

\ifx \showCODEN    \undefined \def \showCODEN     #1{\unskip}     \fi
\ifx \showDOI      \undefined \def \showDOI       #1{#1}\fi
\ifx \showISBNx    \undefined \def \showISBNx     #1{\unskip}     \fi
\ifx \showISBNxiii \undefined \def \showISBNxiii  #1{\unskip}     \fi
\ifx \showISSN     \undefined \def \showISSN      #1{\unskip}     \fi
\ifx \showLCCN     \undefined \def \showLCCN      #1{\unskip}     \fi
\ifx \shownote     \undefined \def \shownote      #1{#1}          \fi
\ifx \showarticletitle \undefined \def \showarticletitle #1{#1}   \fi
\ifx \showURL      \undefined \def \showURL       {\relax}        \fi
% invisible to TeX
\providecommand\bibfield[2]{#2}
\providecommand\bibinfo[2]{#2}
\providecommand\natexlab[1]{#1}
\providecommand\showeprint[2][]{arXiv:#2}

\bibitem[\protect\citeauthoryear{Avrahami, Fried, and Lischinski}{Avrahami
  et~al\mbox{.}}{2022}]%
        {avrahami2022blended_latent}
\bibfield{author}{\bibinfo{person}{Omri Avrahami}, \bibinfo{person}{Ohad
  Fried}, {and} \bibinfo{person}{Dani Lischinski}.}
  \bibinfo{year}{2022}\natexlab{}.
\newblock \showarticletitle{Blended Latent Diffusion}.
\newblock \bibinfo{journal}{{\em arXiv preprint arXiv:2206.02779\/}}
  (\bibinfo{year}{2022}).
\newblock


\bibitem[\protect\citeauthoryear{Baranchuk, Rubachev, Voynov, Khrulkov, and
  Babenko}{Baranchuk et~al\mbox{.}}{2021}]%
        {baranchuk2021labelefficient}
\bibfield{author}{\bibinfo{person}{Dmitry Baranchuk}, \bibinfo{person}{Ivan
  Rubachev}, \bibinfo{person}{Andrey Voynov}, \bibinfo{person}{Valentin
  Khrulkov}, {and} \bibinfo{person}{Artem Babenko}.}
  \bibinfo{year}{2021}\natexlab{}.
\newblock \bibinfo{title}{Label-Efficient Semantic Segmentation with Diffusion
  Models}.
\newblock   (\bibinfo{year}{2021}).
\newblock
\showeprint[arxiv]{cs.CV/2112.03126}


\bibitem[\protect\citeauthoryear{Bock, Tyagi, Kreft, and Alt}{Bock
  et~al\mbox{.}}{2010}]%
        {cell}
\bibfield{author}{\bibinfo{person}{Martin Bock}, \bibinfo{person}{Amit~Kumar
  Tyagi}, \bibinfo{person}{Jan-Ulrich Kreft}, {and} \bibinfo{person}{Wolfgang
  Alt}.} \bibinfo{year}{2010}\natexlab{}.
\newblock \showarticletitle{Generalized voronoi tessellation as a model of
  two-dimensional cell tissue dynamics}.
\newblock \bibinfo{journal}{{\em Bulletin of mathematical biology\/}}
  \bibinfo{volume}{72} (\bibinfo{year}{2010}), \bibinfo{pages}{1696--1731}.
\newblock


\bibitem[\protect\citeauthoryear{Child, Gray, Radford, and Sutskever}{Child
  et~al\mbox{.}}{2019}]%
        {child2019generating}
\bibfield{author}{\bibinfo{person}{Rewon Child}, \bibinfo{person}{Scott Gray},
  \bibinfo{person}{Alec Radford}, {and} \bibinfo{person}{Ilya Sutskever}.}
  \bibinfo{year}{2019}\natexlab{}.
\newblock \showarticletitle{Generating long sequences with sparse
  transformers}.
\newblock \bibinfo{journal}{{\em arXiv preprint arXiv:1904.10509\/}}
  (\bibinfo{year}{2019}).
\newblock


\bibitem[\protect\citeauthoryear{Chung, Fleck, and Forsyth}{Chung
  et~al\mbox{.}}{1998}]%
        {chung1998jigsaw}
\bibfield{author}{\bibinfo{person}{Min~Gyo Chung}, \bibinfo{person}{Margaret~M
  Fleck}, {and} \bibinfo{person}{David~A Forsyth}.}
  \bibinfo{year}{1998}\natexlab{}.
\newblock \showarticletitle{Jigsaw puzzle solver using shape and color}. In
  \bibinfo{booktitle}{{\em ICSP'98. 1998 Fourth International Conference on
  Signal Processing (Cat. No. 98TH8344)}}, Vol.~\bibinfo{volume}{2}. IEEE,
  \bibinfo{pages}{877--880}.
\newblock


\bibitem[\protect\citeauthoryear{Cohen, Sch{\"o}nberger, Speciale, Sattler,
  Frahm, and Pollefeys}{Cohen et~al\mbox{.}}{2016}]%
        {indoor-outdoor-3drec}
\bibfield{author}{\bibinfo{person}{Andrea Cohen}, \bibinfo{person}{Johannes~L.
  Sch{\"o}nberger}, \bibinfo{person}{Pablo Speciale}, \bibinfo{person}{Torsten
  Sattler}, \bibinfo{person}{Jan-Michael Frahm}, {and} \bibinfo{person}{Marc
  Pollefeys}.} \bibinfo{year}{2016}\natexlab{}.
\newblock \showarticletitle{Indoor-Outdoor 3D Reconstruction Alignment}. In
  \bibinfo{booktitle}{{\em Computer Vision -- ECCV 2016}},
  \bibfield{editor}{\bibinfo{person}{Bastian Leibe}, \bibinfo{person}{Jiri
  Matas}, \bibinfo{person}{Nicu Sebe}, {and} \bibinfo{person}{Max Welling}}
  (Eds.). \bibinfo{publisher}{Springer International Publishing},
  \bibinfo{address}{Cham}, \bibinfo{pages}{285--300}.
\newblock


\bibitem[\protect\citeauthoryear{Das, Menon, and Varghese}{Das
  et~al\mbox{.}}{2017}]%
        {shed1}
\bibfield{author}{\bibinfo{person}{Sukhendu Das}, \bibinfo{person}{Arun Menon},
  {and} \bibinfo{person}{Koshy Varghese}.} \bibinfo{year}{2017}\natexlab{}.
\newblock \showarticletitle{Graph-based clustering for apictorial jigsaw
  puzzles of hand shredded content-less pages}.
\newblock \bibinfo{journal}{{\em Lecture Notes in Computer Science (including
  subseries Lecture Notes in Artificial Intelligence and Lecture Notes in
  Bioinformatics)\/}}  \bibinfo{volume}{10127 LNCS} (\bibinfo{year}{2017}),
  \bibinfo{pages}{135 -- 147}.
\newblock


\bibitem[\protect\citeauthoryear{Dhariwal and Nichol}{Dhariwal and
  Nichol}{2021}]%
        {beats}
\bibfield{author}{\bibinfo{person}{Prafulla Dhariwal} {and}
  \bibinfo{person}{Alex Nichol}.} \bibinfo{year}{2021}\natexlab{}.
\newblock \showarticletitle{Diffusion Models Beat GANs on Image Synthesis}.
\newblock \bibinfo{journal}{{\em CoRR\/}}  \bibinfo{volume}{abs/2105.05233}
  (\bibinfo{year}{2021}).
\newblock
\showeprint{2105.05233}
\showURL{%
\url{https://arxiv.org/abs/2105.05233}}


\bibitem[\protect\citeauthoryear{Dosovitskiy, Beyer, Kolesnikov, Weissenborn,
  Zhai, Unterthiner, Dehghani, Minderer, Heigold, Gelly, Uszkoreit, and
  Houlsby}{Dosovitskiy et~al\mbox{.}}{2020}]%
        {vit}
\bibfield{author}{\bibinfo{person}{Alexey Dosovitskiy}, \bibinfo{person}{Lucas
  Beyer}, \bibinfo{person}{Alexander Kolesnikov}, \bibinfo{person}{Dirk
  Weissenborn}, \bibinfo{person}{Xiaohua Zhai}, \bibinfo{person}{Thomas
  Unterthiner}, \bibinfo{person}{Mostafa Dehghani}, \bibinfo{person}{Matthias
  Minderer}, \bibinfo{person}{Georg Heigold}, \bibinfo{person}{Sylvain Gelly},
  \bibinfo{person}{Jakob Uszkoreit}, {and} \bibinfo{person}{Neil Houlsby}.}
  \bibinfo{year}{2020}\natexlab{}.
\newblock \showarticletitle{An Image is Worth 16x16 Words: Transformers for
  Image Recognition at Scale}.
\newblock \bibinfo{journal}{{\em CoRR\/}}  \bibinfo{volume}{abs/2010.11929}
  (\bibinfo{year}{2020}).
\newblock
\showeprint{2010.11929}
\showURL{%
\url{https://arxiv.org/abs/2010.11929}}


\bibitem[\protect\citeauthoryear{Freeman and Garder}{Freeman and
  Garder}{1964}]%
        {freeman1964apictorial}
\bibfield{author}{\bibinfo{person}{Herbert Freeman} {and} \bibinfo{person}{L
  Garder}.} \bibinfo{year}{1964}\natexlab{}.
\newblock \showarticletitle{Apictorial jigsaw puzzles: The computer solution of
  a problem in pattern recognition}.
\newblock \bibinfo{journal}{{\em IEEE Transactions on Electronic Computers\/}}
  (\bibinfo{year}{1964}), \bibinfo{pages}{118--127}.
\newblock


\bibitem[\protect\citeauthoryear{Goldberg, Malon, and Bern}{Goldberg
  et~al\mbox{.}}{2002}]%
        {goldberg2002global}
\bibfield{author}{\bibinfo{person}{David Goldberg},
  \bibinfo{person}{Christopher Malon}, {and} \bibinfo{person}{Marshall Bern}.}
  \bibinfo{year}{2002}\natexlab{}.
\newblock \showarticletitle{A global approach to automatic solution of jigsaw
  puzzles}. In \bibinfo{booktitle}{{\em Proceedings of the eighteenth annual
  symposium on Computational geometry}}. \bibinfo{pages}{82--87}.
\newblock


\bibitem[\protect\citeauthoryear{Gong, Foo, Fan, Ke, Rahmani, and Liu}{Gong
  et~al\mbox{.}}{2023}]%
        {difpose2}
\bibfield{author}{\bibinfo{person}{Jia Gong}, \bibinfo{person}{Lin~Geng Foo},
  \bibinfo{person}{Zhipeng Fan}, \bibinfo{person}{Qiuhong Ke},
  \bibinfo{person}{Hossein Rahmani}, {and} \bibinfo{person}{Jun Liu}.}
  \bibinfo{year}{2023}\natexlab{}.
\newblock \showarticletitle{DiffPose: Toward More Reliable 3D Pose Estimation}.
  In \bibinfo{booktitle}{{\em Proceedings of the IEEE/CVF Conference on
  Computer Vision and Pattern Recognition (CVPR)}}.
\newblock


\bibitem[\protect\citeauthoryear{Guo, Qiu, Liu, Shao, Xue, and Zhang}{Guo
  et~al\mbox{.}}{2019}]%
        {guo2019star}
\bibfield{author}{\bibinfo{person}{Qipeng Guo}, \bibinfo{person}{Xipeng Qiu},
  \bibinfo{person}{Pengfei Liu}, \bibinfo{person}{Yunfan Shao},
  \bibinfo{person}{Xiangyang Xue}, {and} \bibinfo{person}{Zheng Zhang}.}
  \bibinfo{year}{2019}\natexlab{}.
\newblock \showarticletitle{Star-transformer}.
\newblock \bibinfo{journal}{{\em arXiv preprint arXiv:1902.09113\/}}
  (\bibinfo{year}{2019}).
\newblock


\bibitem[\protect\citeauthoryear{Hammoudeh and Pollett}{Hammoudeh and
  Pollett}{2017}]%
        {stitch}
\bibfield{author}{\bibinfo{person}{Zayd Hammoudeh} {and} \bibinfo{person}{Chris
  Pollett}.} \bibinfo{year}{2017}\natexlab{}.
\newblock \showarticletitle{Clustering-Based, Fully Automated Mixed-Bag Jigsaw
  Puzzle Solving}. In \bibinfo{booktitle}{{\em Computer Analysis of Images and
  Patterns}}, \bibfield{editor}{\bibinfo{person}{Michael Felsberg},
  \bibinfo{person}{Anders Heyden}, {and} \bibinfo{person}{Norbert Kr{\"u}ger}}
  (Eds.). \bibinfo{publisher}{Springer International Publishing},
  \bibinfo{address}{Cham}, \bibinfo{pages}{205--217}.
\newblock
\showISBNx{978-3-319-64698-5}


\bibitem[\protect\citeauthoryear{Harel and Ben-Shahar}{Harel and
  Ben-Shahar}{2020}]%
        {harel2020lazy}
\bibfield{author}{\bibinfo{person}{Peleg Harel} {and} \bibinfo{person}{Ohad
  Ben-Shahar}.} \bibinfo{year}{2020}\natexlab{}.
\newblock \showarticletitle{Lazy caterer jigsaw puzzles: Models, properties,
  and a mechanical system-based solver}.
\newblock \bibinfo{journal}{{\em arXiv preprint arXiv:2008.07644\/}}
  (\bibinfo{year}{2020}).
\newblock


\bibitem[\protect\citeauthoryear{Harel and Ben-Shahar}{Harel and
  Ben-Shahar}{2021}]%
        {harel2021crossing}
\bibfield{author}{\bibinfo{person}{Peleg Harel} {and} \bibinfo{person}{Ohad
  Ben-Shahar}.} \bibinfo{year}{2021}\natexlab{}.
\newblock \showarticletitle{Crossing cuts polygonal puzzles: Models and
  Solvers}. In \bibinfo{booktitle}{{\em Proceedings of the IEEE/CVF Conference
  on Computer Vision and Pattern Recognition}}. \bibinfo{pages}{3084--3093}.
\newblock


\bibitem[\protect\citeauthoryear{Ho, Jain, and Abbeel}{Ho
  et~al\mbox{.}}{2020}]%
        {denoising}
\bibfield{author}{\bibinfo{person}{Jonathan Ho}, \bibinfo{person}{Ajay Jain},
  {and} \bibinfo{person}{Pieter Abbeel}.} \bibinfo{year}{2020}\natexlab{}.
\newblock \showarticletitle{Denoising Diffusion Probabilistic Models}.
\newblock \bibinfo{journal}{{\em arXiv preprint arxiv:2006.11239\/}}
  (\bibinfo{year}{2020}).
\newblock


\bibitem[\protect\citeauthoryear{Hoff and Olver}{Hoff and Olver}{2014a}]%
        {puzz1}
\bibfield{author}{\bibinfo{person}{Daniel Hoff} {and} \bibinfo{person}{Peter
  Olver}.} \bibinfo{year}{2014}\natexlab{a}.
\newblock \showarticletitle{Automatic Solution of Jigsaw Puzzles}.
\newblock \bibinfo{journal}{{\em Journal of Mathematical Imaging and Vision\/}}
   \bibinfo{volume}{49} (\bibinfo{date}{05} \bibinfo{year}{2014}).
\newblock
\showDOI{%
\url{https://doi.org/10.1007/s10851-013-0454-3}}


\bibitem[\protect\citeauthoryear{Hoff and Olver}{Hoff and Olver}{2013}]%
        {hoff2013extensions}
\bibfield{author}{\bibinfo{person}{Daniel~J Hoff} {and}
  \bibinfo{person}{Peter~J Olver}.} \bibinfo{year}{2013}\natexlab{}.
\newblock \showarticletitle{Extensions of invariant signatures for object
  recognition}.
\newblock \bibinfo{journal}{{\em Journal of mathematical imaging and vision\/}}
  \bibinfo{volume}{45}, \bibinfo{number}{2} (\bibinfo{year}{2013}),
  \bibinfo{pages}{176--185}.
\newblock


\bibitem[\protect\citeauthoryear{Hoff and Olver}{Hoff and Olver}{2014b}]%
        {hoff2014automatic}
\bibfield{author}{\bibinfo{person}{Daniel~J Hoff} {and}
  \bibinfo{person}{Peter~J Olver}.} \bibinfo{year}{2014}\natexlab{b}.
\newblock \showarticletitle{Automatic solution of jigsaw puzzles}.
\newblock \bibinfo{journal}{{\em Journal of mathematical imaging and vision\/}}
   \bibinfo{volume}{49} (\bibinfo{year}{2014}), \bibinfo{pages}{234--250}.
\newblock


\bibitem[\protect\citeauthoryear{Hosseini and Furukawa}{Hosseini and
  Furukawa}{2023}]%
        {hosseini2023floorplan}
\bibfield{author}{\bibinfo{person}{Sepidehsadat Hosseini} {and}
  \bibinfo{person}{Yasutaka Furukawa}.} \bibinfo{year}{2023}\natexlab{}.
\newblock \bibinfo{title}{Floorplan Restoration by Structure Hallucinating
  Transformer Cascades}.
\newblock   (\bibinfo{year}{2023}).
\newblock
\showeprint[arxiv]{cs.CV/2206.00645}


\bibitem[\protect\citeauthoryear{Kawar, Elad, Ermon, and Song}{Kawar
  et~al\mbox{.}}{2022}]%
        {kawar2022denoising}
\bibfield{author}{\bibinfo{person}{Bahjat Kawar}, \bibinfo{person}{Michael
  Elad}, \bibinfo{person}{Stefano Ermon}, {and} \bibinfo{person}{Jiaming
  Song}.} \bibinfo{year}{2022}\natexlab{}.
\newblock \showarticletitle{Denoising Diffusion Restoration Models}. In
  \bibinfo{booktitle}{{\em Advances in Neural Information Processing Systems}}.
\newblock


\bibitem[\protect\citeauthoryear{Kingma and Ba}{Kingma and Ba}{2014}]%
        {kingma2014adam}
\bibfield{author}{\bibinfo{person}{Diederik~P Kingma} {and}
  \bibinfo{person}{Jimmy Ba}.} \bibinfo{year}{2014}\natexlab{}.
\newblock \showarticletitle{Adam: A method for stochastic optimization}.
\newblock \bibinfo{journal}{{\em arXiv preprint arXiv:1412.6980\/}}
  (\bibinfo{year}{2014}).
\newblock


\bibitem[\protect\citeauthoryear{Kitaev, Kaiser, and Levskaya}{Kitaev
  et~al\mbox{.}}{2020}]%
        {Kitaev2020Reformer:}
\bibfield{author}{\bibinfo{person}{Nikita Kitaev}, \bibinfo{person}{Lukasz
  Kaiser}, {and} \bibinfo{person}{Anselm Levskaya}.}
  \bibinfo{year}{2020}\natexlab{}.
\newblock \showarticletitle{Reformer: The Efficient Transformer}. In
  \bibinfo{booktitle}{{\em International Conference on Learning
  Representations}}.
\newblock
\showURL{%
\url{https://openreview.net/forum?id=rkgNKkHtvB}}


\bibitem[\protect\citeauthoryear{Lambert, Li, Boyadzhiev, Wixson, Narayana,
  Hutchcroft, Hays, Dellaert, and Kang}{Lambert et~al\mbox{.}}{2022a}]%
        {zillow}
\bibfield{author}{\bibinfo{person}{John Lambert}, \bibinfo{person}{Yuguang Li},
  \bibinfo{person}{Ivaylo Boyadzhiev}, \bibinfo{person}{Lambert Wixson},
  \bibinfo{person}{Manjunath Narayana}, \bibinfo{person}{Will Hutchcroft},
  \bibinfo{person}{James Hays}, \bibinfo{person}{Frank Dellaert}, {and}
  \bibinfo{person}{Sing~Bing Kang}.} \bibinfo{year}{2022}\natexlab{a}.
\newblock \showarticletitle{SALVe: Semantic Alignment Verification for
  Floorplan Reconstruction from Sparse Panoramas}. In \bibinfo{booktitle}{{\em
  ECCV}}.
\newblock


\bibitem[\protect\citeauthoryear{Lambert, Li, Boyadzhiev, Wixson, Narayana,
  Hutchcroft, Hays, Dellaert, and Kang}{Lambert et~al\mbox{.}}{2022b}]%
        {zind_sfm}
\bibfield{author}{\bibinfo{person}{John Lambert}, \bibinfo{person}{Yuguang Li},
  \bibinfo{person}{Ivaylo Boyadzhiev}, \bibinfo{person}{Lambert Wixson},
  \bibinfo{person}{Manjunath Narayana}, \bibinfo{person}{Will Hutchcroft},
  \bibinfo{person}{James Hays}, \bibinfo{person}{Frank Dellaert}, {and}
  \bibinfo{person}{Sing~Bing Kang}.} \bibinfo{year}{2022}\natexlab{b}.
\newblock \showarticletitle{SALVe: Semantic Alignment Verification for
  Floorplan Reconstruction from Sparse Panoramas}. In \bibinfo{booktitle}{{\em
  ECCV}}.
\newblock


\bibitem[\protect\citeauthoryear{Le and Li}{Le and Li}{2019}]%
        {le2019jigsawnet}
\bibfield{author}{\bibinfo{person}{Canyu Le} {and} \bibinfo{person}{Xin Li}.}
  \bibinfo{year}{2019}\natexlab{}.
\newblock \showarticletitle{JigsawNet: Shredded image reassembly using
  convolutional neural network and loop-based composition}.
\newblock \bibinfo{journal}{{\em IEEE Transactions on Image Processing\/}}
  \bibinfo{volume}{28}, \bibinfo{number}{8} (\bibinfo{year}{2019}),
  \bibinfo{pages}{4000--4015}.
\newblock


\bibitem[\protect\citeauthoryear{Li, Yang, Chang, Feng, Xu, Li, and Chen}{Li
  et~al\mbox{.}}{2021b}]%
        {SRDiff}
\bibfield{author}{\bibinfo{person}{Haoying Li}, \bibinfo{person}{Yifan Yang},
  \bibinfo{person}{Meng Chang}, \bibinfo{person}{Huajun Feng},
  \bibinfo{person}{Zhihai Xu}, \bibinfo{person}{Qi Li}, {and}
  \bibinfo{person}{Yue{-}ting Chen}.} \bibinfo{year}{2021}\natexlab{b}.
\newblock \showarticletitle{SRDiff: Single Image Super-Resolution with
  Diffusion Probabilistic Models}.
\newblock \bibinfo{journal}{{\em CoRR\/}}  \bibinfo{volume}{abs/2104.14951}
  (\bibinfo{year}{2021}).
\newblock
\showeprint{2104.14951}
\showURL{%
\url{https://arxiv.org/abs/2104.14951}}


\bibitem[\protect\citeauthoryear{Li, Liu, Wang, Liu, and Zeng}{Li
  et~al\mbox{.}}{2021a}]%
        {li2021jigsawgan}
\bibfield{author}{\bibinfo{person}{Ru Li}, \bibinfo{person}{Shuaicheng Liu},
  \bibinfo{person}{Guangfu Wang}, \bibinfo{person}{Guanghui Liu}, {and}
  \bibinfo{person}{Bing Zeng}.} \bibinfo{year}{2021}\natexlab{a}.
\newblock \showarticletitle{Jigsawgan: Auxiliary learning for solving jigsaw
  puzzles with generative adversarial networks}.
\newblock \bibinfo{journal}{{\em IEEE Transactions on Image Processing\/}}
  \bibinfo{volume}{31} (\bibinfo{year}{2021}), \bibinfo{pages}{513--524}.
\newblock


\bibitem[\protect\citeauthoryear{Li, Jin, Xuan, Zhou, Chen, Wang, and Yan}{Li
  et~al\mbox{.}}{2019}]%
        {li2019enhancing}
\bibfield{author}{\bibinfo{person}{Shiyang Li}, \bibinfo{person}{Xiaoyong Jin},
  \bibinfo{person}{Yao Xuan}, \bibinfo{person}{Xiyou Zhou},
  \bibinfo{person}{Wenhu Chen}, \bibinfo{person}{Yu-Xiang Wang}, {and}
  \bibinfo{person}{Xifeng Yan}.} \bibinfo{year}{2019}\natexlab{}.
\newblock \showarticletitle{Enhancing the locality and breaking the memory
  bottleneck of transformer on time series forecasting}.
\newblock \bibinfo{journal}{{\em Advances in neural information processing
  systems\/}}  \bibinfo{volume}{32} (\bibinfo{year}{2019}).
\newblock


\bibitem[\protect\citeauthoryear{Li, Han, Li, and Prisacariu}{Li
  et~al\mbox{.}}{2020}]%
        {li20dualrc}
\bibfield{author}{\bibinfo{person}{Xinghui Li}, \bibinfo{person}{Kai Han},
  \bibinfo{person}{Shuda Li}, {and} \bibinfo{person}{Victor Prisacariu}.}
  \bibinfo{year}{2020}\natexlab{}.
\newblock \showarticletitle{Dual-Resolution Correspondence Networks}. In
  \bibinfo{booktitle}{{\em Conference on Neural Information Processing Systems
  (NeurIPS)}}.
\newblock


\bibitem[\protect\citeauthoryear{Lin, Li, and Wang}{Lin et~al\mbox{.}}{2019}]%
        {lin2019floorplan}
\bibfield{author}{\bibinfo{person}{Cheng Lin}, \bibinfo{person}{Changjian Li},
  {and} \bibinfo{person}{Wenping Wang}.} \bibinfo{year}{2019}\natexlab{}.
\newblock \showarticletitle{Floorplan-jigsaw: Jointly estimating scene layout
  and aligning partial scans}. In \bibinfo{booktitle}{{\em Proceedings of the
  IEEE/CVF International Conference on Computer Vision}}.
  \bibinfo{pages}{5674--5683}.
\newblock


\bibitem[\protect\citeauthoryear{Lin, Maire, Belongie, Hays, Perona, Ramanan,
  Doll{\'a}r, and Zitnick}{Lin et~al\mbox{.}}{2014}]%
        {coco2017}
\bibfield{author}{\bibinfo{person}{Tsung-Yi Lin}, \bibinfo{person}{Michael
  Maire}, \bibinfo{person}{Serge Belongie}, \bibinfo{person}{James Hays},
  \bibinfo{person}{Pietro Perona}, \bibinfo{person}{Deva Ramanan},
  \bibinfo{person}{Piotr Doll{\'a}r}, {and} \bibinfo{person}{C.~Lawrence
  Zitnick}.} \bibinfo{year}{2014}\natexlab{}.
\newblock \showarticletitle{Microsoft COCO: Common Objects in Context}. In
  \bibinfo{booktitle}{{\em Computer Vision -- ECCV 2014}},
  \bibfield{editor}{\bibinfo{person}{David Fleet}, \bibinfo{person}{Tomas
  Pajdla}, \bibinfo{person}{Bernt Schiele}, {and} \bibinfo{person}{Tinne
  Tuytelaars}} (Eds.). \bibinfo{publisher}{Springer International Publishing},
  \bibinfo{address}{Cham}, \bibinfo{pages}{740--755}.
\newblock
\showISBNx{978-3-319-10602-1}


\bibitem[\protect\citeauthoryear{Lin, Liu, Jiang, Do, Tan, and Lu}{Lin
  et~al\mbox{.}}{2016}]%
        {Lin2016RepMatchRF}
\bibfield{author}{\bibinfo{person}{Wen-Yan Lin}, \bibinfo{person}{Siying Liu},
  \bibinfo{person}{Nianjuan Jiang}, \bibinfo{person}{Minh~N. Do},
  \bibinfo{person}{Ping Tan}, {and} \bibinfo{person}{Jiangbo Lu}.}
  \bibinfo{year}{2016}\natexlab{}.
\newblock \showarticletitle{RepMatch: Robust Feature Matching and Pose for
  Reconstructing Modern Cities}. In \bibinfo{booktitle}{{\em ECCV}}.
\newblock


\bibitem[\protect\citeauthoryear{Loshchilov and Hutter}{Loshchilov and
  Hutter}{2017}]%
        {loshchilov2017decoupled}
\bibfield{author}{\bibinfo{person}{Ilya Loshchilov} {and}
  \bibinfo{person}{Frank Hutter}.} \bibinfo{year}{2017}\natexlab{}.
\newblock \showarticletitle{Decoupled weight decay regularization}.
\newblock \bibinfo{journal}{{\em arXiv preprint arXiv:1711.05101\/}}
  (\bibinfo{year}{2017}).
\newblock


\bibitem[\protect\citeauthoryear{Marande and Burger}{Marande and
  Burger}{2007}]%
        {dna2}
\bibfield{author}{\bibinfo{person}{William Marande} {and}
  \bibinfo{person}{Gertraud Burger}.} \bibinfo{year}{2007}\natexlab{}.
\newblock \showarticletitle{Mitochondrial DNA as a Genomic Jigsaw Puzzle}.
\newblock \bibinfo{journal}{{\em Science\/}} \bibinfo{volume}{318},
  \bibinfo{number}{5849} (\bibinfo{year}{2007}), \bibinfo{pages}{415--415}.
\newblock
\showDOI{%
\url{https://doi.org/10.1126/science.1148033}}
\showeprint{https://www.science.org/doi/pdf/10.1126/science.1148033}


\bibitem[\protect\citeauthoryear{Markaki and Panagiotakis}{Markaki and
  Panagiotakis}{2022}]%
        {markaki2022jigsaw}
\bibfield{author}{\bibinfo{person}{Smaragda Markaki} {and}
  \bibinfo{person}{Costas Panagiotakis}.} \bibinfo{year}{2022}\natexlab{}.
\newblock \showarticletitle{Jigsaw puzzle solving techniques and applications:
  a survey}.
\newblock \bibinfo{journal}{{\em The Visual Computer\/}}
  (\bibinfo{year}{2022}), \bibinfo{pages}{1--17}.
\newblock


\bibitem[\protect\citeauthoryear{Martin-Brualla, He, Russell, and
  Seitz}{Martin-Brualla et~al\mbox{.}}{2014}]%
        {martin20143d}
\bibfield{author}{\bibinfo{person}{Ricardo Martin-Brualla},
  \bibinfo{person}{Yanling He}, \bibinfo{person}{Bryan~C Russell}, {and}
  \bibinfo{person}{Steven~M Seitz}.} \bibinfo{year}{2014}\natexlab{}.
\newblock \showarticletitle{The 3d jigsaw puzzle: Mapping large indoor spaces}.
  In \bibinfo{booktitle}{{\em European Conference on Computer Vision}}.
  Springer, \bibinfo{pages}{1--16}.
\newblock


\bibitem[\protect\citeauthoryear{McBride and Kimia}{McBride and Kimia}{2003}]%
        {shed2}
\bibfield{author}{\bibinfo{person}{Jonah~C. McBride} {and}
  \bibinfo{person}{Benjamin~B. Kimia}.} \bibinfo{year}{2003}\natexlab{}.
\newblock \showarticletitle{Archaeological Fragment Reconstruction Using
  Curve-Matching}. In \bibinfo{booktitle}{{\em 2003 Conference on Computer
  Vision and Pattern Recognition Workshop}}, Vol.~\bibinfo{volume}{1}.
  \bibinfo{pages}{3--3}.
\newblock
\showDOI{%
\url{https://doi.org/10.1109/CVPRW.2003.10008}}


\bibitem[\protect\citeauthoryear{Meng, He, Song, Song, Wu, Zhu, and Ermon}{Meng
  et~al\mbox{.}}{2022}]%
        {meng2022sdedit}
\bibfield{author}{\bibinfo{person}{Chenlin Meng}, \bibinfo{person}{Yutong He},
  \bibinfo{person}{Yang Song}, \bibinfo{person}{Jiaming Song},
  \bibinfo{person}{Jiajun Wu}, \bibinfo{person}{Jun-Yan Zhu}, {and}
  \bibinfo{person}{Stefano Ermon}.} \bibinfo{year}{2022}\natexlab{}.
\newblock \showarticletitle{{SDE}dit: Guided Image Synthesis and Editing with
  Stochastic Differential Equations}. In \bibinfo{booktitle}{{\em International
  Conference on Learning Representations}}.
\newblock


\bibitem[\protect\citeauthoryear{Nauata, Hosseini, Chang, Chu, Cheng, and
  Furukawa}{Nauata et~al\mbox{.}}{2021}]%
        {housegan}
\bibfield{author}{\bibinfo{person}{Nelson Nauata},
  \bibinfo{person}{Sepidehsadat Hosseini}, \bibinfo{person}{Kai-Hung Chang},
  \bibinfo{person}{Hang Chu}, \bibinfo{person}{Chin-Yi Cheng}, {and}
  \bibinfo{person}{Yasutaka Furukawa}.} \bibinfo{year}{2021}\natexlab{}.
\newblock \showarticletitle{House-GAN++: Generative Adversarial Layout
  Refinement Network towards Intelligent Computational Agent for Professional
  Architects}. In \bibinfo{booktitle}{{\em Proceedings of the IEEE/CVF
  Conference on Computer Vision and Pattern Recognition}}.
  \bibinfo{pages}{13632--13641}.
\newblock


\bibitem[\protect\citeauthoryear{Nichol and Dhariwal}{Nichol and
  Dhariwal}{2021a}]%
        {impdenoising}
\bibfield{author}{\bibinfo{person}{Alex Nichol} {and} \bibinfo{person}{Prafulla
  Dhariwal}.} \bibinfo{year}{2021}\natexlab{a}.
\newblock \showarticletitle{Improved Denoising Diffusion Probabilistic Models}.
\newblock \bibinfo{journal}{{\em CoRR\/}}  \bibinfo{volume}{abs/2102.09672}
  (\bibinfo{year}{2021}).
\newblock
\showeprint{2102.09672}
\showURL{%
\url{https://arxiv.org/abs/2102.09672}}


\bibitem[\protect\citeauthoryear{Nichol, Dhariwal, Ramesh, Shyam, Mishkin,
  McGrew, Sutskever, and Chen}{Nichol et~al\mbox{.}}{2021}]%
        {gild}
\bibfield{author}{\bibinfo{person}{Alex Nichol}, \bibinfo{person}{Prafulla
  Dhariwal}, \bibinfo{person}{Aditya Ramesh}, \bibinfo{person}{Pranav Shyam},
  \bibinfo{person}{Pamela Mishkin}, \bibinfo{person}{Bob McGrew},
  \bibinfo{person}{Ilya Sutskever}, {and} \bibinfo{person}{Mark Chen}.}
  \bibinfo{year}{2021}\natexlab{}.
\newblock \showarticletitle{{GLIDE:} Towards Photorealistic Image Generation
  and Editing with Text-Guided Diffusion Models}.
\newblock \bibinfo{journal}{{\em CoRR\/}}  \bibinfo{volume}{abs/2112.10741}
  (\bibinfo{year}{2021}).
\newblock
\showeprint{2112.10741}
\showURL{%
\url{https://arxiv.org/abs/2112.10741}}


\bibitem[\protect\citeauthoryear{Nichol and Dhariwal}{Nichol and
  Dhariwal}{2021b}]%
        {cosine}
\bibfield{author}{\bibinfo{person}{Alexander~Quinn Nichol} {and}
  \bibinfo{person}{Prafulla Dhariwal}.} \bibinfo{year}{2021}\natexlab{b}.
\newblock \bibinfo{title}{Improved Denoising Diffusion Probabilistic Models}.
\newblock   (\bibinfo{year}{2021}).
\newblock
\showURL{%
\url{https://openreview.net/forum?id=-NEXDKk8gZ}}


\bibitem[\protect\citeauthoryear{Nielsen, Drewsen, and Hansen}{Nielsen
  et~al\mbox{.}}{2008}]%
        {nielsen2008solving}
\bibfield{author}{\bibinfo{person}{Ture~R Nielsen}, \bibinfo{person}{Peter
  Drewsen}, {and} \bibinfo{person}{Klaus Hansen}.}
  \bibinfo{year}{2008}\natexlab{}.
\newblock \showarticletitle{Solving jigsaw puzzles using image features}.
\newblock \bibinfo{journal}{{\em Pattern Recognition Letters\/}}
  \bibinfo{volume}{29}, \bibinfo{number}{14} (\bibinfo{year}{2008}),
  \bibinfo{pages}{1924--1933}.
\newblock


\bibitem[\protect\citeauthoryear{Noroozi and Favaro}{Noroozi and
  Favaro}{2016}]%
        {noroozi2016unsupervised}
\bibfield{author}{\bibinfo{person}{Mehdi Noroozi} {and} \bibinfo{person}{Paolo
  Favaro}.} \bibinfo{year}{2016}\natexlab{}.
\newblock \showarticletitle{Unsupervised learning of visual representations by
  solving jigsaw puzzles}. In \bibinfo{booktitle}{{\em Computer Vision--ECCV
  2016: 14th European Conference, Amsterdam, The Netherlands, October 11-14,
  2016, Proceedings, Part VI}}. Springer, \bibinfo{pages}{69--84}.
\newblock


\bibitem[\protect\citeauthoryear{Paszke, Gross, Massa, Lerer, Bradbury, Chanan,
  Killeen, Lin, Gimelshein, Antiga, et~al\mbox{.}}{Paszke
  et~al\mbox{.}}{2019}]%
        {paszke2019pytorch}
\bibfield{author}{\bibinfo{person}{Adam Paszke}, \bibinfo{person}{Sam Gross},
  \bibinfo{person}{Francisco Massa}, \bibinfo{person}{Adam Lerer},
  \bibinfo{person}{James Bradbury}, \bibinfo{person}{Gregory Chanan},
  \bibinfo{person}{Trevor Killeen}, \bibinfo{person}{Zeming Lin},
  \bibinfo{person}{Natalia Gimelshein}, \bibinfo{person}{Luca Antiga},
  {et~al\mbox{.}}} \bibinfo{year}{2019}\natexlab{}.
\newblock \showarticletitle{Pytorch: An imperative style, high-performance deep
  learning library}.
\newblock \bibinfo{journal}{{\em Advances in neural information processing
  systems\/}}  \bibinfo{volume}{32} (\bibinfo{year}{2019}).
\newblock


\bibitem[\protect\citeauthoryear{Pop}{Pop}{2009}]%
        {dna}
\bibfield{author}{\bibinfo{person}{Mihai Pop}.}
  \bibinfo{year}{2009}\natexlab{}.
\newblock \showarticletitle{{Genome assembly reborn: recent computational
  challenges}}.
\newblock \bibinfo{journal}{{\em Briefings in Bioinformatics\/}}
  \bibinfo{volume}{10}, \bibinfo{number}{4} (\bibinfo{date}{05}
  \bibinfo{year}{2009}), \bibinfo{pages}{354--366}.
\newblock
\showISSN{1467-5463}
\showDOI{%
\url{https://doi.org/10.1093/bib/bbp026}}
\showeprint{https://academic.oup.com/bib/article-pdf/10/4/354/847444/bbp026.pdf}


\bibitem[\protect\citeauthoryear{Radack and Badler}{Radack and Badler}{1982}]%
        {radack1982jigsaw}
\bibfield{author}{\bibinfo{person}{Gerald~M Radack} {and}
  \bibinfo{person}{Norman~I Badler}.} \bibinfo{year}{1982}\natexlab{}.
\newblock \showarticletitle{Jigsaw puzzle matching using a boundary-centered
  polar encoding}.
\newblock \bibinfo{journal}{{\em Computer Graphics and Image Processing\/}}
  \bibinfo{volume}{19}, \bibinfo{number}{1} (\bibinfo{year}{1982}),
  \bibinfo{pages}{1--17}.
\newblock


\bibitem[\protect\citeauthoryear{Ramesh, Dhariwal, Nichol, Chu, and
  Chen}{Ramesh et~al\mbox{.}}{2022}]%
        {clip}
\bibfield{author}{\bibinfo{person}{Aditya Ramesh}, \bibinfo{person}{Prafulla
  Dhariwal}, \bibinfo{person}{Alex Nichol}, \bibinfo{person}{Casey Chu}, {and}
  \bibinfo{person}{Mark Chen}.} \bibinfo{year}{2022}\natexlab{}.
\newblock \showarticletitle{Hierarchical text-conditional image generation with
  clip latents}.
\newblock \bibinfo{journal}{{\em arXiv preprint arXiv:2204.06125\/}}
  (\bibinfo{year}{2022}).
\newblock


\bibitem[\protect\citeauthoryear{Rombach, Blattmann, Lorenz, Esser, and
  Ommer}{Rombach et~al\mbox{.}}{2021}]%
        {rombach2021highresolution}
\bibfield{author}{\bibinfo{person}{Robin Rombach}, \bibinfo{person}{Andreas
  Blattmann}, \bibinfo{person}{Dominik Lorenz}, \bibinfo{person}{Patrick
  Esser}, {and} \bibinfo{person}{Björn Ommer}.}
  \bibinfo{year}{2021}\natexlab{}.
\newblock \bibinfo{title}{High-Resolution Image Synthesis with Latent Diffusion
  Models}.
\newblock   (\bibinfo{year}{2021}).
\newblock
\showeprint[arxiv]{cs.CV/2112.10752}


\bibitem[\protect\citeauthoryear{Saharia, Ho, Chan, Salimans, Fleet, and
  Norouzi}{Saharia et~al\mbox{.}}{2021}]%
        {saharia2021}
\bibfield{author}{\bibinfo{person}{Chitwan Saharia}, \bibinfo{person}{Jonathan
  Ho}, \bibinfo{person}{William Chan}, \bibinfo{person}{Tim Salimans},
  \bibinfo{person}{David~J Fleet}, {and} \bibinfo{person}{Mohammad Norouzi}.}
  \bibinfo{year}{2021}\natexlab{}.
\newblock \showarticletitle{Image super-resolution via iterative refinement}.
\newblock \bibinfo{journal}{{\em arXiv:2104.07636\/}} (\bibinfo{year}{2021}).
\newblock


\bibitem[\protect\citeauthoryear{Sarlin, DeTone, Malisiewicz, and
  Rabinovich}{Sarlin et~al\mbox{.}}{2020}]%
        {arlin20superglue}
\bibfield{author}{\bibinfo{person}{Paul-Edouard Sarlin},
  \bibinfo{person}{Daniel DeTone}, \bibinfo{person}{Tomasz Malisiewicz}, {and}
  \bibinfo{person}{Andrew Rabinovich}.} \bibinfo{year}{2020}\natexlab{}.
\newblock \showarticletitle{{SuperGlue}: Learning Feature Matching with Graph
  Neural Networks}. In \bibinfo{booktitle}{{\em CVPR}}.
\newblock
\showURL{%
\url{https://arxiv.org/abs/1911.11763}}


\bibitem[\protect\citeauthoryear{Sasaki, Willcocks, and Breckon}{Sasaki
  et~al\mbox{.}}{2021}]%
        {trans}
\bibfield{author}{\bibinfo{person}{Hiroshi Sasaki}, \bibinfo{person}{Chris~G.
  Willcocks}, {and} \bibinfo{person}{Toby~P. Breckon}.}
  \bibinfo{year}{2021}\natexlab{}.
\newblock \showarticletitle{{UNIT-DDPM:} UNpaired Image Translation with
  Denoising Diffusion Probabilistic Models}.
\newblock \bibinfo{journal}{{\em CoRR\/}}  \bibinfo{volume}{abs/2104.05358}
  (\bibinfo{year}{2021}).
\newblock
\showeprint{2104.05358}
\showURL{%
\url{https://arxiv.org/abs/2104.05358}}


\bibitem[\protect\citeauthoryear{Shabani, Hosseini, and Furukawa}{Shabani
  et~al\mbox{.}}{2022}]%
        {shabani2022housediffusion}
\bibfield{author}{\bibinfo{person}{Mohammad~Amin Shabani},
  \bibinfo{person}{Sepidehsadat Hosseini}, {and} \bibinfo{person}{Yasutaka
  Furukawa}.} \bibinfo{year}{2022}\natexlab{}.
\newblock \showarticletitle{HouseDiffusion: Vector Floorplan Generation via a
  Diffusion Model with Discrete and Continuous Denoising}.
\newblock \bibinfo{journal}{{\em arXiv preprint arXiv:2211.13287\/}}
  (\bibinfo{year}{2022}).
\newblock


\bibitem[\protect\citeauthoryear{Shabani, Song, Odamaki, Fujiki, and
  Furukawa}{Shabani et~al\mbox{.}}{2021}]%
        {amin}
\bibfield{author}{\bibinfo{person}{Mohammad~Amin Shabani},
  \bibinfo{person}{Weilian Song}, \bibinfo{person}{Makoto Odamaki},
  \bibinfo{person}{Hirochika Fujiki}, {and} \bibinfo{person}{Yasutaka
  Furukawa}.} \bibinfo{year}{2021}\natexlab{}.
\newblock \showarticletitle{Extreme Structure from Motion for Indoor Panoramas
  without Visual Overlaps}. In \bibinfo{booktitle}{{\em Proceedings of the
  IEEE/CVF International Conference on Computer Vision (ICCV)}}.
\newblock
\showURL{%
\url{https://aminshabani.github.io/publications/extreme_sfm/pdfs/iccv2021_2088.pdf}}


\bibitem[\protect\citeauthoryear{Shan, Liu, Zhang, Wang, Han, Wang, Ma, and
  Gao}{Shan et~al\mbox{.}}{2023}]%
        {diffpose}
\bibfield{author}{\bibinfo{person}{Wenkang Shan}, \bibinfo{person}{Zhenhua
  Liu}, \bibinfo{person}{Xinfeng Zhang}, \bibinfo{person}{Zhao Wang},
  \bibinfo{person}{Kai Han}, \bibinfo{person}{Shanshe Wang},
  \bibinfo{person}{Siwei Ma}, {and} \bibinfo{person}{Wen Gao}.}
  \bibinfo{year}{2023}\natexlab{}.
\newblock \showarticletitle{Diffusion-Based 3D Human Pose Estimation with
  Multi-Hypothesis Aggregation}.
\newblock \bibinfo{journal}{{\em arXiv preprint arXiv:2303.11579\/}}
  (\bibinfo{year}{2023}).
\newblock


\bibitem[\protect\citeauthoryear{Shih and Lu}{Shih and Lu}{2018}]%
        {shih2018divide}
\bibfield{author}{\bibinfo{person}{Huang-Chia Shih} {and}
  \bibinfo{person}{Chien-Liang Lu}.} \bibinfo{year}{2018}\natexlab{}.
\newblock \showarticletitle{Divide-and-conquer jigsaw puzzle solving}. In
  \bibinfo{booktitle}{{\em 2018 IEEE Visual Communications and Image Processing
  (VCIP)}}. IEEE, \bibinfo{pages}{1--2}.
\newblock


\bibitem[\protect\citeauthoryear{Snavely, Seitz, and Szeliski}{Snavely
  et~al\mbox{.}}{2006}]%
        {snavely2006photo}
\bibfield{author}{\bibinfo{person}{Noah Snavely}, \bibinfo{person}{Steven~M
  Seitz}, {and} \bibinfo{person}{Richard Szeliski}.}
  \bibinfo{year}{2006}\natexlab{}.
\newblock \showarticletitle{Photo tourism: exploring photo collections in 3D}.
\newblock \bibinfo{journal}{{\em ACM siggraph\/}} (\bibinfo{year}{2006}),
  \bibinfo{pages}{835--846}.
\newblock


\bibitem[\protect\citeauthoryear{Song, Sohl{-}Dickstein, Kingma, Kumar, Ermon,
  and Poole}{Song et~al\mbox{.}}{2020}]%
        {color}
\bibfield{author}{\bibinfo{person}{Yang Song}, \bibinfo{person}{Jascha
  Sohl{-}Dickstein}, \bibinfo{person}{Diederik~P. Kingma},
  \bibinfo{person}{Abhishek Kumar}, \bibinfo{person}{Stefano Ermon}, {and}
  \bibinfo{person}{Ben Poole}.} \bibinfo{year}{2020}\natexlab{}.
\newblock \showarticletitle{Score-Based Generative Modeling through Stochastic
  Differential Equations}.
\newblock \bibinfo{journal}{{\em CoRR\/}}  \bibinfo{volume}{abs/2011.13456}
  (\bibinfo{year}{2020}).
\newblock
\showeprint{2011.13456}
\showURL{%
\url{https://arxiv.org/abs/2011.13456}}


\bibitem[\protect\citeauthoryear{Sun, Shen, Wang, Bao, and Zhou}{Sun
  et~al\mbox{.}}{2021}]%
        {sun2021loftr}
\bibfield{author}{\bibinfo{person}{Jiaming Sun}, \bibinfo{person}{Zehong Shen},
  \bibinfo{person}{Yuang Wang}, \bibinfo{person}{Hujun Bao}, {and}
  \bibinfo{person}{Xiaowei Zhou}.} \bibinfo{year}{2021}\natexlab{}.
\newblock \showarticletitle{{LoFTR}: Detector-Free Local Feature Matching with
  Transformers}.
\newblock \bibinfo{journal}{{\em {CVPR}\/}} (\bibinfo{year}{2021}).
\newblock


\bibitem[\protect\citeauthoryear{Toler-Franklin, Brown, Weyrich, Funkhouser,
  and Rusinkiewicz}{Toler-Franklin et~al\mbox{.}}{2010}]%
        {toler2010multi}
\bibfield{author}{\bibinfo{person}{Corey Toler-Franklin},
  \bibinfo{person}{Benedict Brown}, \bibinfo{person}{Tim Weyrich},
  \bibinfo{person}{Thomas Funkhouser}, {and} \bibinfo{person}{Szymon
  Rusinkiewicz}.} \bibinfo{year}{2010}\natexlab{}.
\newblock \showarticletitle{Multi-feature matching of fresco fragments}.
\newblock \bibinfo{journal}{{\em ACM Transactions on Graphics (TOG)\/}}
  \bibinfo{volume}{29}, \bibinfo{number}{6} (\bibinfo{year}{2010}),
  \bibinfo{pages}{1--12}.
\newblock


\bibitem[\protect\citeauthoryear{Wei, Ding, Park, Sajnani, Poulenard, Sridhar,
  and Guibas}{Wei et~al\mbox{.}}{2023}]%
        {legonet}
\bibfield{author}{\bibinfo{person}{Qiuhong~Anna Wei}, \bibinfo{person}{Sijie
  Ding}, \bibinfo{person}{Jeong~Joon Park}, \bibinfo{person}{Rahul Sajnani},
  \bibinfo{person}{Adrien Poulenard}, \bibinfo{person}{Srinath Sridhar}, {and}
  \bibinfo{person}{Leonidas Guibas}.} \bibinfo{year}{2023}\natexlab{}.
\newblock \showarticletitle{LEGO-Net: Learning Regular Rearrangements of
  Objects in Rooms}.
\newblock \bibinfo{journal}{{\em arXiv preprint arXiv:2301.09629\/}}
  (\bibinfo{year}{2023}).
\newblock


\bibitem[\protect\citeauthoryear{Wolfson, Schonberg, Kalvin, and
  Lamdan}{Wolfson et~al\mbox{.}}{1988}]%
        {wolfson1988solving}
\bibfield{author}{\bibinfo{person}{Haim Wolfson}, \bibinfo{person}{Edith
  Schonberg}, \bibinfo{person}{Alan Kalvin}, {and} \bibinfo{person}{Yehezkel
  Lamdan}.} \bibinfo{year}{1988}\natexlab{}.
\newblock \showarticletitle{Solving jigsaw puzzles by computer}.
\newblock \bibinfo{journal}{{\em Annals of Operations Research\/}}
  \bibinfo{volume}{12}, \bibinfo{number}{1} (\bibinfo{year}{1988}),
  \bibinfo{pages}{51--64}.
\newblock


\bibitem[\protect\citeauthoryear{Wolleb, Sandkühler, Bieder, Valmaggia, and
  Cattin}{Wolleb et~al\mbox{.}}{2021}]%
        {wolle}
\bibfield{author}{\bibinfo{person}{Julia Wolleb}, \bibinfo{person}{Robin
  Sandkühler}, \bibinfo{person}{Florentin Bieder}, \bibinfo{person}{Philippe
  Valmaggia}, {and} \bibinfo{person}{Philippe~C. Cattin}.}
  \bibinfo{year}{2021}\natexlab{}.
\newblock \bibinfo{title}{Diffusion Models for Implicit Image Segmentation
  Ensembles}.
\newblock   (\bibinfo{year}{2021}).
\newblock
\showeprint[arxiv]{cs.CV/2112.03145}


\bibitem[\protect\citeauthoryear{Wu, Fu, Tang, Wang, Qi, and Liu}{Wu
  et~al\mbox{.}}{2019}]%
        {wu2019data}
\bibfield{author}{\bibinfo{person}{Wenming Wu}, \bibinfo{person}{Xiao-Ming Fu},
  \bibinfo{person}{Rui Tang}, \bibinfo{person}{Yuhan Wang},
  \bibinfo{person}{Yu-Hao Qi}, {and} \bibinfo{person}{Ligang Liu}.}
  \bibinfo{year}{2019}\natexlab{}.
\newblock \showarticletitle{Data-driven interior plan generation for
  residential buildings}.
\newblock \bibinfo{journal}{{\em ACM Transactions on Graphics (TOG)\/}}
  \bibinfo{volume}{38}, \bibinfo{number}{6} (\bibinfo{year}{2019}),
  \bibinfo{pages}{1--12}.
\newblock


\bibitem[\protect\citeauthoryear{Yang, Pan, Luo, Zhou, Grauman, and Huang}{Yang
  et~al\mbox{.}}{2019}]%
        {Yang_2019_CVPR}
\bibfield{author}{\bibinfo{person}{Zhenpei Yang}, \bibinfo{person}{Jeffrey~Z.
  Pan}, \bibinfo{person}{Linjie Luo}, \bibinfo{person}{Xiaowei Zhou},
  \bibinfo{person}{Kristen Grauman}, {and} \bibinfo{person}{Qixing Huang}.}
  \bibinfo{year}{2019}\natexlab{}.
\newblock \showarticletitle{Extreme Relative Pose Estimation for RGB-D Scans
  via Scene Completion}. In \bibinfo{booktitle}{{\em Proceedings of the
  IEEE/CVF Conference on Computer Vision and Pattern Recognition (CVPR)}}.
\newblock


\bibitem[\protect\citeauthoryear{Yang, Yan, and Huang}{Yang
  et~al\mbox{.}}{2020}]%
        {yang2020extreme}
\bibfield{author}{\bibinfo{person}{Zhenpei Yang}, \bibinfo{person}{Siming Yan},
  {and} \bibinfo{person}{Qixing Huang}.} \bibinfo{year}{2020}\natexlab{}.
\newblock \showarticletitle{Extreme relative pose network under hybrid
  representations}. In \bibinfo{booktitle}{{\em Proceedings of the IEEE/CVF
  Conference on Computer Vision and Pattern Recognition}}.
  \bibinfo{pages}{2455--2464}.
\newblock


\bibitem[\protect\citeauthoryear{Yi, Trulls, Ono, Lepetit, Salzmann, and
  Fua}{Yi et~al\mbox{.}}{2018}]%
        {yi2018learning}
\bibfield{author}{\bibinfo{person}{Kwang~Moo Yi}, \bibinfo{person}{Eduard
  Trulls}, \bibinfo{person}{Yuki Ono}, \bibinfo{person}{Vincent Lepetit},
  \bibinfo{person}{Mathieu Salzmann}, {and} \bibinfo{person}{Pascal Fua}.}
  \bibinfo{year}{2018}\natexlab{}.
\newblock \showarticletitle{Learning to find good correspondences}. In
  \bibinfo{booktitle}{{\em Proceedings of the IEEE conference on computer
  vision and pattern recognition}}. \bibinfo{pages}{2666--2674}.
\newblock


\bibitem[\protect\citeauthoryear{Zhao, Bao, Li, and Zhu}{Zhao
  et~al\mbox{.}}{2022}]%
        {trans2}
\bibfield{author}{\bibinfo{person}{Min Zhao}, \bibinfo{person}{Fan Bao},
  \bibinfo{person}{Chongxuan Li}, {and} \bibinfo{person}{Jun Zhu}.}
  \bibinfo{year}{2022}\natexlab{}.
\newblock \showarticletitle{Egsde: Unpaired image-to-image translation via
  energy-guided stochastic differential equations}.
\newblock \bibinfo{journal}{{\em arXiv preprint arXiv:2207.06635\/}}
  (\bibinfo{year}{2022}).
\newblock


\end{thebibliography}
